%% file: root.tex
\documentclass[letterpaper, 10 pt, journal, twoside]{IEEEtran}  % Comment this line out if you need a4paper

\IEEEoverridecommandlockouts                              % This command is only needed if 
\usepackage{amsmath} % assumes amsmath package installed
\usepackage{amssymb}  % assumes amsmath package installed
\usepackage{graphicx} % for pdf, bitmapped graphics files

\usepackage{booktabs}
\let\labelindent\relax
\usepackage{enumitem}

\usepackage[breaklinks,colorlinks]{hyperref}
\usepackage{siunitx}
\usepackage[ruled,vlined]{algorithm2e}
\usepackage{algorithmic}
\usepackage{microtype}
\usepackage{flushend}
\usepackage[hang,flushmargin]{footmisc}

\usepackage{url}
\usepackage[table, dvipsnames]{xcolor}

\makeatletter
\g@addto@macro{\endtabular}{\rowfont{}}% Clear row font
\makeatother
\newcommand{\rowfonttype}{}% Current row font
\newcommand{\rowfont}[1]{% Set current row font
\gdef\rowfonttype{#1}#1\ignorespaces%
}
\makeatother
\usepackage[export]{adjustbox}

\usepackage{pifont}% http://ctan.org/pkg/pifont
\newcommand{\cmark}{\ding{51}}%
\usepackage{siunitx}
\usepackage{cuted}
\usepackage{threeparttable}
\usepackage{multirow}
\usepackage{cite}

\usepackage{tikz}
\usepackage{pgfplots}

\usepackage[resetlabels]{multibib}
\newcites{S}{References}

\usepackage{tabularx}
\newcolumntype{Y}{>{\centering\arraybackslash}X}
\newcolumntype{Z}{>{\raggedleft\arraybackslash}X}

\definecolor{dark-green}{RGB}{12,80,12}

\newcommand{\secref}[1]{Sec.~\ref{#1}}
\renewcommand{\eqref}[1]{Eq.~(\ref{#1})}
\newcommand{\figref}[1]{Fig.~\ref{#1}}
\newcommand{\tabref}[1]{Tab.~\ref{#1}}

\newcommand{\appref}[1]{Supplementary Sec.~\ref{#1}}

\newcommand{\para}[1]{\parskip=5pt\noindent\textit{#1}}
\newcommand{\ours}{MoMa-LLM}
\newcommand{\ourslong}{MoMa-LLM}
\newcommand{\website}{\url{http://moma-llm.cs.uni-freiburg.de}}

\newcommand{\myworries}[1]{\textcolor{black}{#1}}
\newcommand{\myworriestable}{\color{black}} 

\renewcommand{\baselinestretch}{0.985}

\title{\LARGE \bf
Language-Grounded Dynamic Scene Graphs for\\Interactive Object Search with Mobile Manipulation
}

\author{Daniel Honerkamp$^{1*}$, Martin Büchner$^{1*}$, Fabien Despinoy$^{2}$, Tim Welschehold$^{1}$, Abhinav Valada$^{1}$% <-this % stops a space
\thanks{$^{*}$ Equal contribution.}%
\thanks{$^{1}$ Department of Computer Science, University of Freiburg, Germany.}%
\thanks{$^{2}$ Toyota Motor Europe (TME).}%
\thanks{This work was funded by Toyota Motor Europe (TME) and an academic grant from NVIDIA. We thank Aron Distelzweig for his help in implementing the HIMOS model for this task.}% <-this % stops a space
\thanks{© 2024 IEEE.  Personal use of this material is permitted.  Permission from IEEE must be obtained for all other uses, in any current or future media, including reprinting/republishing this material for advertising or promotional purposes, creating new collective works, for resale or redistribution to servers or lists, or reuse of any copyrighted component of this work in other works.}%
}

\begin{document}

\maketitle
% \thispagestyle{empty}
% \pagestyle{empty}

%%%%%%%%%%%%%%%%%%%%%%%%%%%%%%%%%%%%%%%%%%%%%%%%%%%%%%%%%%%%%%%%%%%%%%%%%%%%%%%%
\input{sections/0_abstract}
% Keywords appear just beneath the abstract. Use only for final RAL version.

\input{sections/1_introduction}

\input{sections/2_related_work}
\input{sections/3_approach}
\input{sections/4_experiments}

\vspace{-0.2em}
\input{sections/5_conclusion}

%%%%%%%%%%%%%%%%%%%%%%%%%%%%%%%%%%%%%%%%%%%%%%%%%%%%%%%%%%%%%%%%%%%%%%%%%%%%%%%%

\footnotesize
\vspace{-0.3em}
\bibliographystyle{IEEEtran}
\bibliography{bibliography.bib}

% \appendix
\input{sections/6_appendix}

%%%%%%%%%%%%%%%%%%%%%%%%%%%%%%%%%%%%%%%%%%%%%%%%%%%%%%%%%%%%%%%%%%%%%%%%%%%%%%%%

\end{document}

%% file: sections/0_abstract.tex
\begin{abstract}
To fully leverage the capabilities of mobile manipulation robots, it is imperative that they are able to autonomously execute long-horizon tasks in large unexplored environments. While large language models (LLMs) have shown emergent reasoning skills on arbitrary tasks, existing work primarily concentrates on explored environments, typically focusing on either navigation or manipulation tasks in isolation. In this work, we propose \ours{}, a novel approach that grounds language models within structured representations derived from open-vocabulary scene graphs, dynamically updated as the environment is explored. We tightly interleave these representations with an object-centric action space. \myworries{Given object detections,} the resulting approach is zero-shot, open-vocabulary, and readily extendable to a spectrum of mobile manipulation and household robotic tasks. We demonstrate the effectiveness of \ours{} in a novel semantic interactive search task in large realistic indoor environments. In extensive experiments in both simulation and the real world, we show substantially improved search efficiency compared to conventional baselines and state-of-the-art approaches, as well as its applicability to more abstract tasks. We make the code publicly available at \website{}.
\end{abstract}

%% file: sections/1_introduction.tex
\section{Introduction}
\IEEEPARstart{I}{nteractive} embodied AI tasks in large, unexplored, human-centered environments require reasoning over long horizons and a multitude of objects. In many cases, the considered environments are a priori unknown or continuously rearranged.
Recent advancements have demonstrated the potential of large language models (LLMs) in generating high-level plans~\cite{huang2023voxposer, liu2023reflect, li2023interactive, lin2023text2motion}. However, these efforts have predominantly focused on fully observed environments such as table-top manipulation, or a priori explored scenes, struggling to generate executable and grounded plans suitable for real-world robotic execution. This problem is strongly exacerbated in large scenes with numerous objects and long time horizons. In turn, this increases the risk of generating impractical sequences or hallucinations~\cite{rana2023sayplan, agia2022taskography}. Furthermore, the presence of interactive scenes and articulated objects introduces a multitude of potential states and failure cases. To address these challenges, we propose grounding LLMs in dynamically built scene graphs. Our approach incorporates a scene understanding module that\myworries{, given object detections,} constructs open-vocabulary scene graphs from dense maps and Voronoi graphs. These diverse representations are then tightly interweaved with an object-centric action space. Leveraging the current scene representation, we extract structured and compact textual representations of the scene to facilitate efficient planning with pre-trained LLMs.\looseness=-1

\input{figures/tex-figures/teaser}

To evaluate the efficacy of our approach, we formulate an interactive semantic search task, extending previous non-semantic interactive tasks~\cite{schmalstieg2023learning} to more complex scenarios. In this task, an agent has to find a target object within an indoor environment, encapsulating real-world challenges such as opening doors to navigate through the environment, and searching inside cabinets and drawers to find the desired object. This task is challenging as it requires reasoning about manipulation and navigation skills, operating in unexplored environments, spanning large apartments with numerous rooms and objects. Consequently, it is representative of more complex mobile manipulation tasks while retaining the specificity required for thorough evaluations and comparisons against conventional methods. Furthermore, we introduce a novel evaluation paradigm for object search tasks, employing full efficiency curves to remove the dependency on arbitrary time budgets inherent in existing methods. Additionally, we propose the \textit{AUC-E} metric to distill these curves into a single metric for coherent evaluation. We perform extensive experimental evaluations in both simulation and the real-world, and demonstrate that given appropriately structured representations, LLMs can leverage their accumulated knowledge about the human world to achieve exceptional results, outperforming state-of-the-art approaches across diverse fields. Our approach is zero-shot, with open-vocabulary \myworries{reasoning}, and inherently scalable to various mobile manipulation and household robotic tasks, as we demonstrate on a set of abstract search tasks.

To summarize, our main contributions are
\begin{itemize}[topsep=0pt]
    \item A scalable scene representation centered around a dynamic scene graph with open-vocabulary room clustering and classification.
    \item Structured compact knowledge extraction to ground LLMs in scene graphs for large unexplored environments.
    \item Semantic interactive search task for large scenes with numerous objects and receptacles.
    \item Novel evaluation paradigm for object search tasks through full efficiency curves, instead of a single time budget.
    \item We release the code at \website{}.
\end{itemize}

%% file: figures/tex-figures/teaser.tex
\setlength{\tabcolsep}{1pt}

\begin{figure}[t]
	\centering
	\resizebox{.85\linewidth}{!}{%
 \includegraphics[width=\linewidth,trim={0cm 0cm 0cm 0cm},clip,angle =0,valign=c]{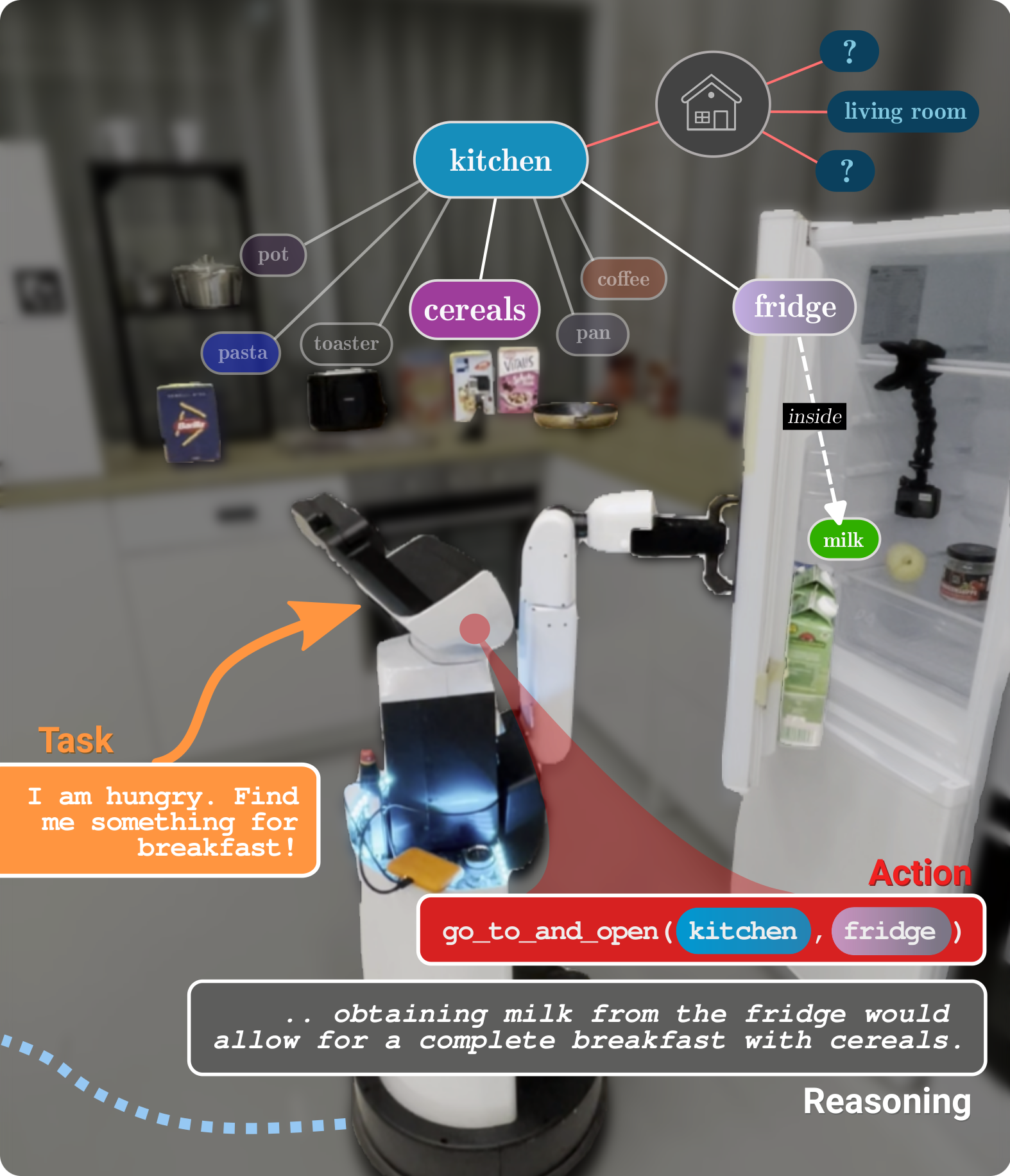}}

     \caption{\ours{} performs long-horizon interactive object search in household environments from language queries using dynamically built scene graphs.\looseness=-1} 
  	\label{fig:teaser}
\vspace{-0.6cm}
\end{figure}
\setlength{\tabcolsep}{6pt}

%% file: sections/2_related_work.tex
\section{Related Work}

{\parskip=0pt
\noindent\textit{3D Scene Graphs}
serve as sparse environment representations that abstract from dense semantic maps~\cite{werby23hovsg}. The disassembly of large scenes into objects, regions, etc., and their representation as nodes thus provides hierarchical and object-centric representations. In addition, nodes and edges may contain semantic attributes~\cite{hughes2022hydra, armeni20193d, greve2023curb, wu_scenegraphfusion}. 
Despite the lower geometric fidelity compared to dense maps, scene graphs prove particularly successful in the realm of high-level reasoning and planning, while providing a powerful interface with mapping~\cite{greve2023curb, rana2023sayplan}. 
Orthogonally, Hydra~\cite{hughes2022hydra} focuses on representing dynamically changing scenes.
Multiple works have explored the use of scene graphs for reasoning in mobile robotics. Most build a hierarchical scene graph of the form (building, floors, rooms, objects). Different variants add edges among objects~\cite{gu2023conceptgraphs}, add a Voronoi graph~\cite{wu2024voronav} for storing observations, or separate the set of objects into static and potentially moving objects~\cite{rajvanshi2023saynav}. While ConceptGraphs~\cite{gu2023conceptgraphs} and VoroNav~\cite{wu2024voronav} investigate the use of zero-shot perception inputs for task planning, others such as SayPlan~\cite{rana2023sayplan}, SayNav~\cite{rajvanshi2023saynav}, and Taskography~\cite{agia2022taskography} focus on the reasoning task itself by utilizing ground truth semantic scene graphs~\cite{ni2023grid, chalvatzaki2023learning}. 

Realizing object navigation using both dynamic and interactive scene graphs has not been tackled thus far in the aforementioned works.}

{\parskip=3pt
\noindent\textit{Language Models for Planning}: Several recent works have investigated language models' abilities to generate task plans for robotic manipulation. These largely focus either on static table-top scenes of limited size and a limited number of objects~\cite{huang2023voxposer, li2023interactive, lin2023text2motion}, or a fully observable scene.
On the other side of the spectrum, vision-language-navigation investigates pure navigation tasks in large apartments to either navigate along a described path or towards a specific instance of an object~\cite{gu2022vision, song2023llm}. 
%NavGPT~\cite{song2023llm} extracts language descriptions of the panoramic views around the robot with BLIP2, then uses an LLM to aggregate them into a single language description of the local surrounding of the robot for a vision-and-language navigation task.
A smaller number of works have investigated apartment-wide mobile manipulation tasks. LLM-Planner~\cite{song2023llm} uses information retrieval of the closest matching known task. Chalvatzaki~\textit{et~al.}~\cite{chalvatzaki2023learning} finetune an LLM to encode object-object relations extracted from a scene graph.
SayCan~\cite{ichter2022do} combines affordance values with language scores.
However, all of these methods focus on tasks restricted to single rooms.
SayPlan~\cite{rana2023sayplan} focuses on identifying relevant subgraphs in large known scene graphs by iteratively extracting or collapsing nodes. Ni~\textit{et~al.}~\cite{ni2023grid} learn a transformer-based model on top of a frozen LLM to predict subtasks from fully known scene graphs. 
In contrast, we focus on interactive search in large fully unexplored environments.
As a result, simple prompting strategies, such as lists of observed objects~\cite{song2023llm, gu2023conceptgraphs, ni2023grid} or raw JSON input~\cite{rana2023sayplan} of a full scene graph to a language model becomes insufficient, as we demonstrate in our experiments.} 

{\parskip=3pt
\noindent\textit{Object Search} has been tackled via a wide range of methods, including classical methods such as frontier exploration~\cite{yamauchi1997frontier}, vision-based reinforcement learning~\cite{chaplot2020learning}, or auditory signals~\cite{younes2023catch}. Graph Neural networks (GNNs) have been used with scene graphs to find specific object instances with hierarchical and relational constraints~\cite{lingelbach2023task} or in frequently changing, dynamic scenes~\cite{kurenkov2023modeling, ying2023rp}.
Schmalstieg~\textit{et~al.}~\cite{schmalstieg2023learning} introduced the interactive search task, in which an agent has to open doors and search through cabinets and drawers. While they focus on random target placements and a restricted number of objects and receptacles, we introduce a semantic single-object search variation of this task, which uses all objects in the scene and keeps the semantic co-occurrences in the scene intact.}

Non-interactive semantic search has been previously tackled. Most recent methods used language models to extract similarities or co-occurrences with the target object to score frontiers~\cite{zhou2023esc, chen2023not, yokoyama2023vlfm} or predict potential functions towards a target object with supervised learning~\cite{ramakrishnan2022poni}. While these works focus on pairwise score calculations, we treat it as a planning problem in which the full scene is encoded jointly.
In contrast to these works, we consider objects that are not freely accessible and require interaction with the environment and thereby reasoning over multiple steps such as opening doors and receptacles instead of pure directional reasoning.

Lastly, \myworries{given object detections,} our \myworries{representation and reasoning} is fully open-vocabulary - both in terms of room and object categories.
Conceptually most similar to our work, SayNav~\cite{rajvanshi2023saynav} utilizes a scene graph together with an LLM. However, it focuses on non-interactive search, restricting the LLM's access to a room subgraph, assumes restrictions such as knowledge about scene graph edges, and relies on a hardcoded heuristic of when to go to the next (already open) door.\looseness=-1 

%% file: sections/3_approach.tex
\section{Problem Statement: Embodied Reasoning}%

\input{figures/tex-figures/overview}
In our setting, an embodied, robotic agent is situated in a large, unexplored environment and has to complete a given task, described by a language goal $g$. The agent is acting in a Partially Observable Markov Decision Process (POMDP) $\mathcal{M} = (\mathcal{S}, \mathcal{A}, \mathcal{O}, T(s' | s, a), P(o | s), r(s, a))$ where $\mathcal{S}, \mathcal{A}$ and $\mathcal{O}$ are the state, action and observation spaces, $T$ and $P$ describe the transition and observation probabilities, $s$, $s'$ are the \myworries{underlying} current and next state, $o$ is the \myworries{agent's} current observation \myworries{consisting of posed RGB-D frame $I_t$}, $a$ is the current action and $r$ is the reward. To succeed in these tasks, the agent has to perceive the environment and create a representation while reasoning about how to complete the tasks through exploration and interaction with the environment. 

We introduce the task of \textit{semantic interactive object search}. In contrast to most existing works~\cite{chaplot2020learning, zhou2023esc, chen2023not, schmalstieg2022learning}, interactive object search requires manipulation of the environment to navigate and explore it. As in realistic, human-centric environments, doors may block pathways and objects are not openly visible but may be stored away in receptacles like drawers or cabinets.
We extend the interactive task introduced in~\cite{schmalstieg2023learning} to a much larger number of objects and receptacles and a prior distribution of realistic room-object and object-object relations. As a result, other objects in the scene can provide valuable information about the position of the target. 
While existing tasks such as the Habitat challenge and Robothor use semantic placements, they do not support any physical interactions or objects placed within receptacles. 

We implement the task in the iGibson scenes~\cite{li2021igibson}, consisting of 15 interactive apartments based on scans of real houses. At the beginning of an episode, all doors are closed and the agent is given a task description in natural language. The task is deemed successful if the agent has observed an instance of the target category and calls $done()$.

The iGibson scenes contain realistic furniture and room distributions, but few other objects are placed in relation to this. We enrich the scenes with realistic object placements, both within receptacles and on top of furniture, by extending and matching previously introduced prior distributions $P^{prior}$ over room and object relations~\cite{kurenkov2023modeling}\myworries{, by aligning room names manually and matching object names via SBERT cosine similarities. We then assume that all objects that can be found on top of an object and that fit in size,
can also be found inside it and vice-versa}. 
Given a valid scene instantiation, we then draw a target category $g \sim U(scene)$ from all categories in the scene.
This results in the procedural generation of a wide range of tasks over 84 possible target classes. 
Further details can be found in the \appref{app:environment}.\looseness=-1

\section{\ourslong{}}
To address the challenges of interactive open-vocabulary household tasks, we propose \textit{\ours{}}, which intertwines high-level reasoning with scalable dynamic scene representations. We ground large-language models in hierarchical 3D scene graphs $\mathcal{G}_{S}$ that hold object- and room-level entities as well as a more fine-grained Voronoi graph for navigation. The LLM provides high-level actions that are executed through low-level skills as shown in \figref{fig:scene-llm}. In general, we assume access to ground truth perception \myworries{for semantic masks, depth, localization and handle detection} as the focus of this work is on the reasoning aspect.

\subsection{Hierarchical 3D Scene Graph}

To provide an LLM with structured input, we craft a hierarchical scene graph that includes a navigational Voronoi graph.\looseness=-1

\subsubsection{Dynamic RGB-D Mapping}
The agent perceives posed RGB-D frames $\{I_{0}, \dots, I_{t}\}$ including semantics from the environment. The contained points are transformed into the global coordinate frame and arranged on a 3D voxel grid $\mathcal{M}_{t}$. As we tackle an interactive problem, our map is dynamically updated based on novel explored areas or the occurrence of object dynamics in the scene. To infer obstacle positions, walls and explored free space, we first obtain the highest occupied entry per stixel in $\mathcal{M}_{t}$. These entries are then turned into a two-dimensional bird's-eye-view (BEV) occupancy map $\mathcal{B}_{t}$ by inferring all occupied positions except for those classified as free space $\mathcal{F}_{t}$. The latter in turn represents the navigable area that is used for robot exploration.

\subsubsection{Voronoi Graph} \myworries{Similar to Hydra~\cite{hughes2022hydra},} we abstract from the created dense maps by computing a navigational graph $\mathcal{G}_{\mathcal{V}}$. 
%that is used in downstream tasks to associate objects in close vicinity or estimate geodesic distances. %By evaluating whether stixels of $\mathcal{M}_{t}$ hold the \textit{wall}, \textit{door} or \textit{window} semantic category, we construct a wall map $\mathcal{W}_{t}$. To increase robustness, 
We first inflate $\mathcal{B}_{t}$ using an Euclidean signed distance field (ESDF) formulation for robustness, but overwrite free space coordinates as given in $\mathcal{F}_{t}$ as zero. Based on this, we compute a Generalized Voronoi Diagram (GVD) that holds a set of points $\mathcal{V}$ with the same clearance to the closest obstacles drawn from $\mathcal{B}_{t}$. 

We exclude all nodes that lie in the immediate vicinity of obstacles or do not reside within $\mathcal{B}_{t}$. Given the GVD boundaries, we  construct edges $\mathcal{E}$ among $\mathcal{V}$ and obtain our navigational Voronoi graph $\mathcal{G}_{\mathcal{V}} = (\mathcal{V}, \mathcal{E})$. Throughout our experiments, we found that extracting the largest connected component of the graph provides the robot-centric Voronoi graph while other components commonly lie outside the explored area. Lastly, we sparsify $\mathcal{G}_{\mathcal{V}}$ to obtain fewer navigational nodes. 

\subsubsection{3D Scene Graph}\label{sec:scenegraph}
The \ours-policy operates on an attributed 3D scene graph $\mathcal{G}_{S}$ that holds different abstraction levels, namely rooms and objects. We first separate the global Voronoi graph $\mathcal{G}_{\mathcal{V}}$ into multiple regions. To do so, we eliminate edges and nodes of $G_{\mathcal{V}}$ near doors \myworries{instead of separating graphs at geometrical constrictions~\cite{hughes2022hydra}}. Using a mixture of Gaussians, we generate a two-dimensional probability distribution over all observed door \myworries{positions} in the environment:
\begin{equation}
    % \footnotesize
    \rho_{\mathcal{N}}(\boldsymbol{x}, \boldsymbol{H}) = \frac{1}{N_{D}}\sum_{i=1}^{N_D} K_{\boldsymbol{H}}(\boldsymbol{x}-\boldsymbol{x_i}),
\end{equation}
where $\boldsymbol{x}_{i}= (x_i,y_i)$ are the door center coordinates, $K_{\boldsymbol{H}}$ is the scaled Gaussian kernel of observed doors and $\boldsymbol{H}$ the bandwidth matrix\myworries{, which we set to 2.0 based on manual tuning on the training scenes}. %Edges that fall into these regions are identified by computing the discrete integral of probability values along the edge. 
Edges \myworries{that fall into high-probability regions and} exceed an empirically tuned probability threshold are disregarded along with isolated nodes. Following this principle, we obtain the separated Voronoi graph $\mathcal{G}_{\mathcal{V}}^{R}$ covering distinct rooms.
In the next step, we infer the high-level connectivity among rooms by calculating the shortest paths between nodes of $\mathcal{G}_{\mathcal{V}}$ that belong to disjoint components of $\mathcal{G}_{\mathcal{V}}^{R}$. Whenever a path \textit{traverses} just two distinct rooms as given by $\mathcal{G}_{\mathcal{V}}^{R}$, the two rooms count as immediate neighbors. 
% Based on this connectivity, our policy is provided with adjacent room information given the current robot position.
Finally, we map objects to rooms. For each object $o \in \mathcal{G}_{S}$, we identify the node that minimizes the distance \myworries{$d_{vo}$} to the closest viewpoint $v_p$ from which the object was seen. To this end, we calculate the shortest path from the object $o$ to this viewpoint. It consists of the path on the Voronoi graph \myworries{$\mathcal{G}_{\mathcal{V}}$,} and the Euclidean distances $d$ from the Voronoi nodes \myworries{$n_o$ and $n_{v_p}$} to the object $o$ and viewpoint $v_p$, respectively. By weighting the distance to the object with an exponent of $\lambda=1.3$, we ensure to prefer nodes close to the object. \myworries{Objects are then assigned to the room label $R$ of the node $n_o$ that minimizes \eqref{eq:object_assignment}}. This prohibits the erroneous assignments of objects to a neighboring room through walls. Doors may be connected to multiple rooms. 
\begin{equation}\label{eq:object_assignment}
\myworries{d_{vo}} = \min_{n_o \myworries{, n_{v_p}} \in \mathcal{G}_{\mathcal{V}}^{R}} path(n_o, n_{v_p}) + d(o, n_o)^{\lambda} + d(v_p, n_{v_p})
\end{equation}

\input{figures/tex-figures/room-cls}

\subsubsection{Room Classification}
Similar to Chen~\textit{et~al.}~\cite{chen2022leveraging}, we perform room classification by providing an LLM with the set of object categories contained in each room. We perform this as open-set classification, in which we let the LLM freely pick the room categories deemed most appropriate. 

The resulting LLM prompts are detailed in \figref{fig:roomprompt}. Room classification is performed in each high-level policy step, as the explored scene and scene graph evolve.
\myworries{We provide a concise overview of all scene graph layers in \tabref{tab:scene_graph_structure}.}

\subsection{High-Level Action Space}

We design an object-centric action space, which is tightly intertwined with the different granularities of the scene representation. It consists of the following high-level actions:

    \para{navigate(\texttt{room\_name}, \texttt{object\_name}):} Navigation to an object in a room via an A$^*$ planner in the explored BEV-map $\mathcal{B}_{t}$, inflated by \SI{0.1}{\meter}. It first navigates to the Voronoi node associated with the object, then to the most central, free point on an arc around the object. This enables robust navigation to objects in partially explored space and ensures navigation to the correct room through the Voronoi assignment detailed in \secref{sec:scenegraph}. Navigation is considered successful if the agent reaches within \SI{1.5}{\meter} of the object.\\
    \para{go\_to\_and\_open(\texttt{room\_name},  \texttt{object\_name}):} Navigate to a specific object, then open it. For doors, continue to navigate into the opened door frame.\\ 
    \para{close(\texttt{room\_name},  \texttt{object\_name}):} Equivalent to opening.\\
    \para{explore(\texttt{room\_name}):} Move to an unexplored frontier within this room. Deemed successful if within \SI{0.5}{\meter} of the frontier.\\
    \para{done():} Terminate the episode and evaluate if the target object has been found.

Ambiguities of multiple instances of the specified class in a room are resolved by selecting the closest instance. 
The subpolicies then generate actions in the low-level action space and return once they succeed or encounter a failure. Throughout their execution, they continuously update the scene representations. Refer to the \appref{app:environment} for details.\looseness=-1

\vspace{-0.2cm}
\subsection{Grounded High-Level Planning}
\input{figures/tex-figures/policy}

We encode the accumulated knowledge of the scene graph into natural language by extracting the relevant components and embedding them in a problem-specific structured manner. Our method fulfills three properties: (i) grounding - guiding the LLM to adhere to the physical realities of the scene, (ii) specificity - avoiding long or irrelevant context queries that increase hallucinations and the difficulty of the planning problem~\cite{rana2023sayplan, agia2022taskography}, and (iii) open-set - our \myworries{reasoning} is open-vocabulary and performs in a zero-shot manner, enabling direct deployment with unknown semantics and perception models. The resulting prompt for the language model is shown in \figref{fig:prompt}. In the following, we describe the main components of structured encoding. We demonstrate the importance of this structure in \secref{sec:experiments}. \looseness=-1

\subsubsection{Scene Structure} 
We encode the main room-object structure from the scene graph into a structured list of rooms and their containing objects and encode path distances (based on an A$^*$-planner) by binning them and mapping them to adjectives~\cite{chalvatzaki2023learning}, as detailed in \appref{app:language}.
We then employ the following filtering to allow for compact text encodings: we summarize matching nodes within a room with a counter, we filter out open doors that provide no new connectivity, and we encode object states directly within the object name, e.g. as \textit{"opened"} or \textit{"closed [object-name]"}.

\subsubsection{Partial Observability}
As the environment is initially unknown, it requires explicit reasoning about exploration-exploitation trade-offs. We identify frontiers to explorable areas~\cite{yamauchi1997frontier}, then leverage the scene graph to provide them with semantic meaning. Firstly, we associate each frontier with a room through matching with $G_{\mathcal{V}}^{R}$. Secondly, we apply hole-filling to the BEV map to differentiate whether a frontier is an encapsulated area within a room, such as occluded space behind furniture, or whether the frontier is leading out to new areas.
Correspondingly, we then represent them as \textit{"unexplored area"} within a room, while frontiers that lead to other areas are listed separately, see \figref{fig:prompt}. 
The second type of unexplored space is receptacles that may contain target objects. Together with the encoded object states, we find that the language model is capable of inferring affordances from the object descriptions, removing the need to explicitly encode them. If trying to open objects that cannot be opened, the according subpolicy will fail and the LLM has to reason about an appropriate response.

\subsubsection{History in Dynamic Scenes}
Given the size of the scenes, the conversation history quickly grows too large to provide to a language model directly.
Instead, we aim to find the most compact representation of previous actions to fulfill the Markov property. For each high-level decision, we encode the latest scene representation and start a new query to the LLM. As the scene representation is dynamically updated, this automatically encodes all newly acquired knowledge.
To account for previous interactions, we provide the LLM with a history of the last $h$ actions. But as the scene graph changes dynamically, the previous room- and object-centric function calls may no longer match the current scene. Instead, we keep track of interaction positions, and then re-align the previous actions by matching the positions to their closest Voronoi nodes and associated room labels.
We then provide the LLM with a list of the re-aligned function calls, as shown in \figref{fig:prompt}. \myworries{E.g., the agent executes \textit{explore(\texttt{living room})}. But revealing a fridge, later classifies the same room as kitchen. The realigned history will then correctly reflect this action as \textit{explore(\texttt{kitchen})}.}\looseness=-1

\subsubsection{Re-trial and Re-planning}\label{sec:replanning}
Extracting meaningful feedback for failure reasons for robots in the real world remains an open problem~\cite{liu2023reflect}, as the number of possible failure reasons is almost unlimited. Instead, we provide very limited feedback about subpolicy success, which can be readily generated in the real world. We rely on a simple success state to the action history, stating \textit{"success", "failure"}, or \textit{"invalid argument"} in case the output of the LLM could not be matched to the scene graph.
We differentiate two cases of replanning: if the agent attempted interactions or commands that cannot be parsed or are deemed infeasible without attempting execution, we have not gained any new information about the scene, and we continue the conversation with the message \textit{"The last action $<$function-call$>$ failed. Please try another command."}. In case of more than five failures without state change, we terminate the episode as unsuccessful.
If a subpolicy attempted execution but failed to complete its task, we re-encode the latest scene, update the action history, and let the LLM make a normal next decision with the updated state.

%% file: figures/tex-figures/overview.tex
\setlength{\tabcolsep}{1pt}
\begin{figure*}[!ht]
	\centering
	\resizebox{\textwidth}{!}{%
 \includegraphics[width=\textwidth,trim={0cm 0cm 0cm 0cm},clip,angle =0,valign=c]{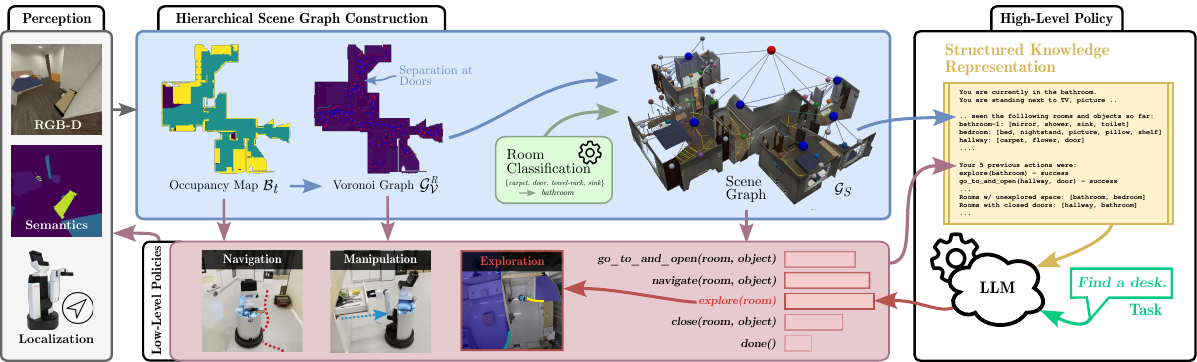}}

     \caption{\ours{}: From posed RGB-D images and semantics, we construct a semantic 3D map from which we extract a various occupancy maps in the BEV space and construct a navigational Voronoi graph. Through room clustering and room-object assigments we then build up a hierarchical scene graph. From this scalable scene representation, we extract the task-relevant knowledge and encode it into a structured language representation. A large language model then produces high-level commands which are executed by low-level subpolicies. These in turn draw on and update the scene representations.}
  	\label{fig:scene-llm}
   \vspace{-0.3cm}
\end{figure*}
\setlength{\tabcolsep}{6pt}

%% file: figures/tex-figures/room-cls.tex
\setlength{\tabcolsep}{1pt}

\begin{figure}[t]
	\centering
	\resizebox{1.0\linewidth}{!}{%
 \includegraphics[width=1.0\textwidth,trim={0cm 0cm 0cm 0cm},clip,angle =0,valign=c]
 {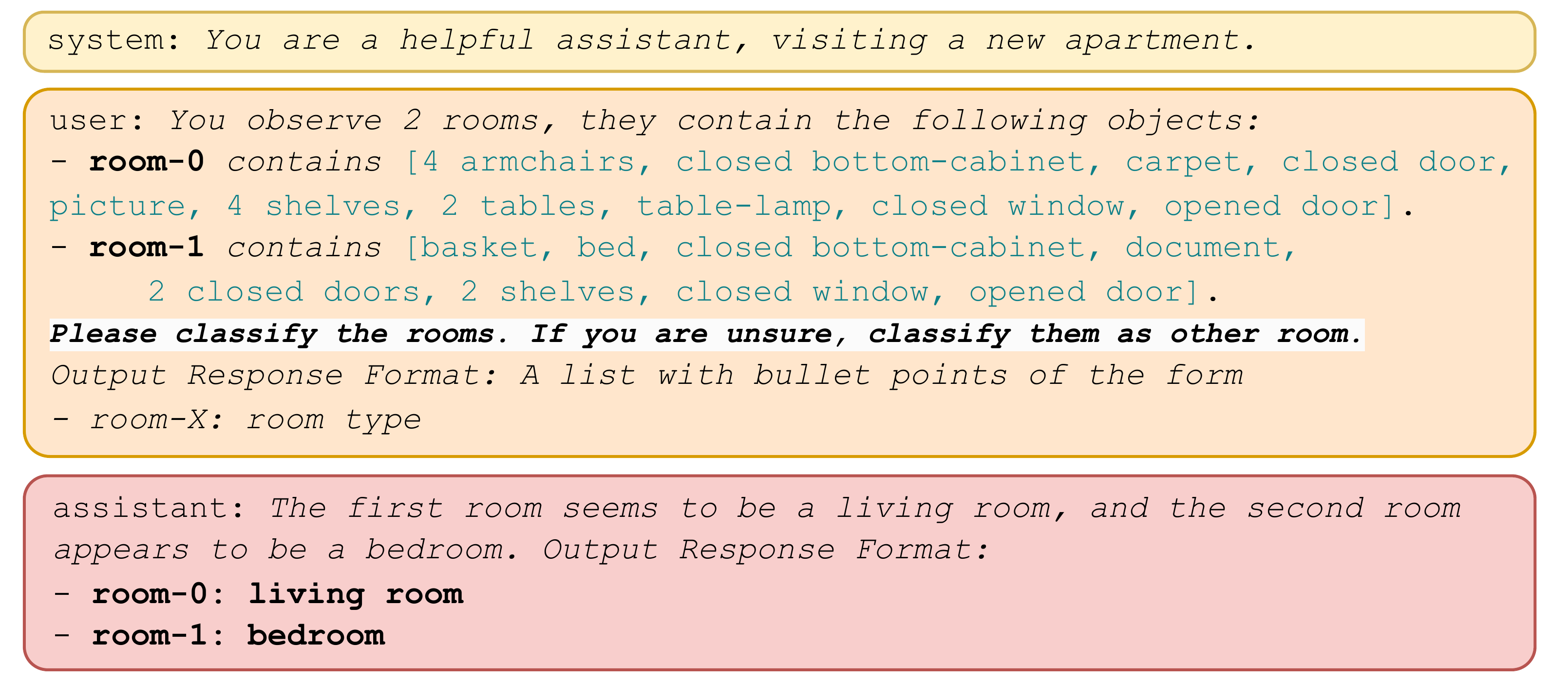}}
     \caption{Room Classification Prompt: based on the objects and room clusters of the scene graph, an LLM performs open-vocabulary classification.}
  	\label{fig:roomprompt}
   \vspace{-0.4cm}
\end{figure} 
\setlength{\tabcolsep}{6pt}

%% file: figures/tex-figures/policy.tex
\setlength{\tabcolsep}{1pt}
\begin{figure*}[ht]
	\centering
	\resizebox{.90\linewidth}{!}{%
 \includegraphics[width=\textwidth,trim={0cm 0cm 0cm 0cm},clip,angle =0,valign=c]{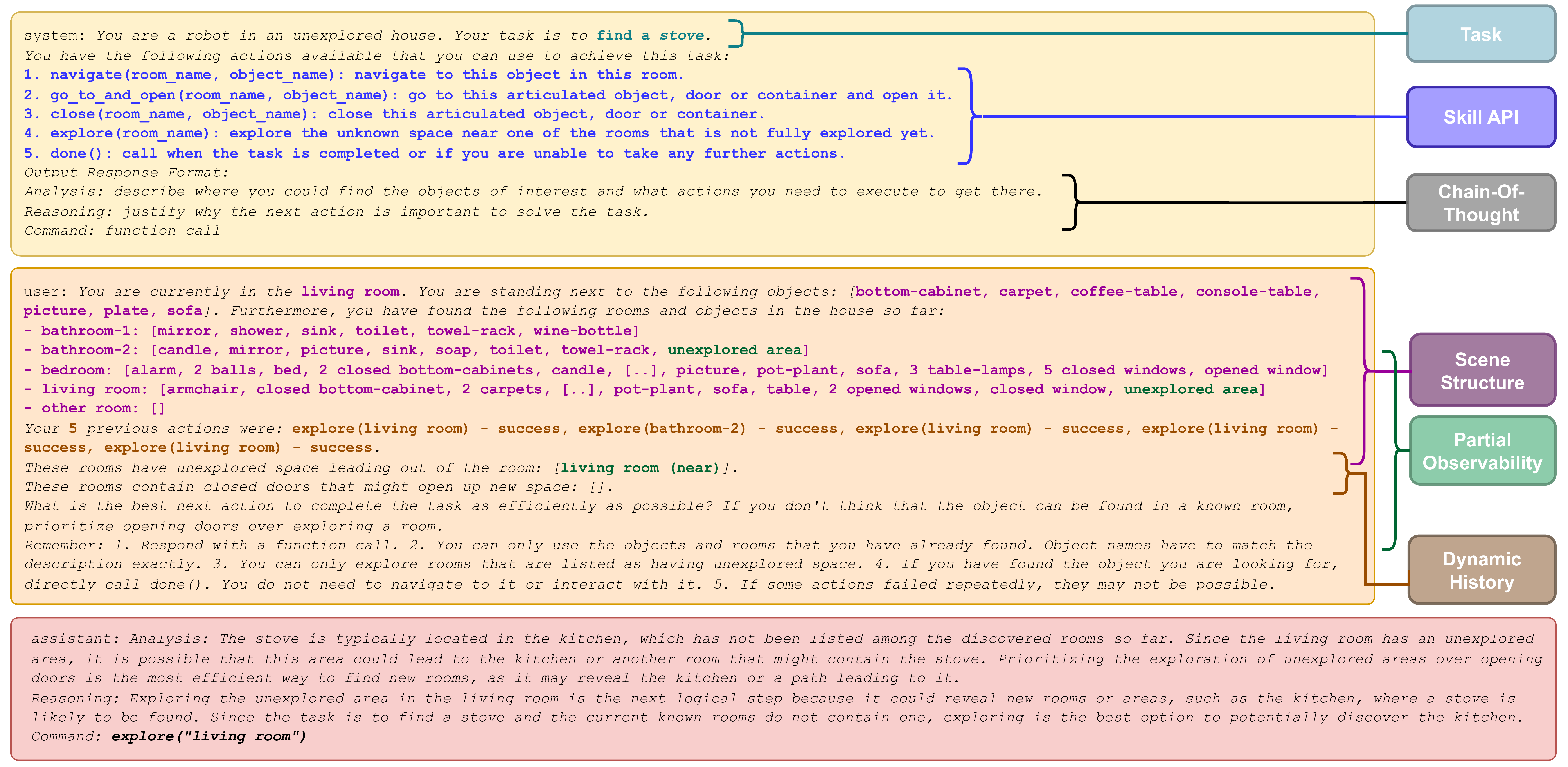}}

     \caption{High-level Reasoning Prompt: We encode the extracted scene representation to natural language, providing structured information to a language model.}
  	\label{fig:prompt}
   \vspace{-0.3cm}
\end{figure*}
\setlength{\tabcolsep}{6pt}

%% file: sections/4_experiments.tex
\section{Experiments}\label{sec:experiments}
As language models, we use \textsl{gpt-4-1106-preview} for the high-level reasoning and \textsl{gpt-3.5-turbo-1106} for the room classification task~\cite{openai2023gpt}. \myworries{For simplicity, we recompute the scene graph each time step. More advanced implementations would reduce costs through incremental updates.}

\para{Baselines:}
We compare our approach against heuristic-based, recent learning-based, and language-based methods. We provide all baselines except Unstructured LLM with a ground truth $done()$ decision when the object has been observed.\\
\para{Random:} uniform random choice among all available actions \myworries{(detected frontiers and closed objects)}. \\
\para{Greedy:} greedily triggers the closest \myworries{available} action \myworries{based on the shortest path calculated by an A$^*$-planner}.\\

\para{ESC-Interact\myworries{ive}:} ESC is a recent approach for semantic object search~\cite{zhou2023esc} which scores frontiers based on object-object and object-room co-occurrences as well as their distance. We extend the approach to interactive search by using the same rules to score openable objects and then select the action with the highest value. 
Co-occurrences are based on similarities of a finetuned Deberta-v3 language model~\cite{he2022debertav3}, following the authors' instructions. To isolate the impact of the decision making, we use the same scene graph and low-level policies as for our method.\\
\para{HIMOS:} A hierarchical reinforcement learning approach~\cite{schmalstieg2023learning} which learns to combine frozen low-level policies for interactive object search, based on a semantic map memory. We adapt it by giving it the same subpolicies as our approach and scale it to the much larger number of objects in our scenes by restricting the instance navigation to target and articulated objects.\looseness=-1\\
\para{Unstructured LLM:} This baseline provides the scene graph in a JSON format without any additional structure to the language model. The prompt structure is derived from SayPlan~\cite{rana2023sayplan}, adapted to the instructions and scene graph of our method. See \appref{app:baselines} for a prompt example.\looseness=-1\\
\myworries{\para{\ours{} w/ Hydra:} We incorporate the room segmentation approach introduced by Hydra~\cite{hughes2022hydra} into our scene graph construction pipeline to measure the impact of our proposed door-wise room separation mechanism.}

\para{Metrics:} We use three types of metrics to evaluate methods.\\
\para{Success rate (SR):} the share of episodes in which the agent finds the target object. 
We terminate an episode if the agent reaches 50 high-level steps, indicating being stuck.\\
\para{Success weighted by Path Length (SPL)~\cite{anderson2018evaluation}} calculates the fraction of distance traveled to the shortest possible path and weights it by whether the episode was successful. This metric does not take into account the costs of object interactions.\\
\para{Search efficiency curve and AUC-E:} While the commonly used success and SPL metrics allow for reducing the evaluation to a single number, they rely on an arbitrarily set maximum allowed time budget or number of environment steps. 
As a result, these metrics do not differentiate between methods that search thoroughly but on average take more steps versus methods that search large areas very quickly, but that might fail to search every corner for small objects. Instead, we reason that the desired time budget depends heavily on the use case and propose a new metric that evaluates the full efficiency curve. For each possible budget (number of steps), we calculate the share of episodes that succeeded with this or fewer number of steps. This results in an efficiency curve, in which the best policies are located in the top left corner, enabling the comparison of success rates for arbitrary budgets.
We can still reduce this to a single number by calculating the area under the efficiency curve (\textit{AUC-E}). A perfect (but unachievable) policy, that finds all objects in a single step will have a value of one, a policy that does not find any objects will have a value of zero. We calculate the integral up to 5,000 low-level steps, at which points almost all methods make no further progress.\looseness=-1

\input{tables/sim_experiments}

\input{figures/tex-figures/auc_curve_pgf}

\subsection{Simulation Experiments}
We instantiate the task in the iGibson simulator~\cite{li2021igibson} with a Fetch robot. In contrast to previous LLM-based works~\cite{rana2023sayplan, rajvanshi2023saynav}, we evaluate all approaches in completely unseen apartments, following the data split of the iGibson challenge into eight training scenes for the development of all modules and prompt engineering and seven test scenes. For each scene, we evaluate the agent over 25 procedurally generated episodes with randomized start poses, target objects, and object distributions.

\para{Scene Understanding:}
\myworries{We compare our door-based room separation algorithm against Hydra~\cite{hughes2022hydra}, which separates a Voronoi graph of places based on dilating obstacles. We evaluate the room segmentation precision and recall as defined by Hughes~\textit{et~al.}~\cite{hughes2022hydra} and the separated Voronoi graph purity at all high-level policy steps to account for robustness throughout exploration. As depicted in Tab.~\ref{tab:igibson_env_partitioning}, we observe greater average precision and recall of MoMa-LLM in terms of dense region segmentation as well as a lower variance across time. The sparsely evaluated graph purity shows that the separated Voronoi graphs of MoMa-LLM cover fewer ground-truth rooms per predicted room than graphs produced by Hydra, which may produce inferior results when facing non-apparent constrictions or object clutter. We found that real-world scenes contained in iGibson regularly feature constant-diameter corridors and narrow passages due to furniture placements, which impede detecting rooms based on geometrical constrictions. This demonstrates that room separation algorithms benefit from semantic cues such as detected doors, door frames, archways, or changing floor materials at room boundaries.} We found our policy to be robust to under-segmented rooms even though objects from multiple rooms were, e.g., considered part of a single room. By relying on the camera pose from which an object is observed we reduce the number of wrong object-room assignments \textit{through} walls. Following the door-wise separation of rooms, our approach however is prone to \textit{open} room concepts such as combined kitchen and living rooms. For more information and graph depictions, refer to the \appref{app:scene_graph_acc}.\looseness=-1

\input{tables/env_partitioning}

\para{Policies:}
The results and efficiency curves for the search task are shown in \tabref{tab:sim_experiments} and \figref{fig:auc-vis-pgf}.
We find that, given appropriate subpolicies, heuristics can complete a significant share of episodes. However, they are not sufficient for an efficient search strategy, resulting in low SPL and AUC-E. 
Similarly, while HIMOS achieves a high success rate, it is unable to explore efficiently. We found that the RL agent struggled with the much larger action space that resulted from the many more interactable instances in our scenes.
ESC in contrast, is able to exploit the co-occurrences to improve over the other baselines. However, given its pair-wise comparisons, it is unable to optimize over longer action sequences. In contrast, \ours{} achieves similar success rates as HIMOS with a much higher search efficiency, both in terms of SPL and AUC-E. We find that the structured prompt representation is essential for this, with the Unstructured LLM performing much worse.
\myworries{We then perform a number of ablations of the language encodings. We find that encoding the frontiers is very important. Removing the history also leads to a, although smaller, drop in performance. Even a coarser representation of the history, consisting of only a list of visited rooms similar to \cite{rana2023sayplan}, is already beneficial, but slightly worse than the full action history. Lastly, we evaluate the impact of not encoding distances nor nearby objects and also find a small drop in performance.}

This picture is fortified by the full efficiency curves in \figref{fig:auc-vis-pgf}, which show that the \ours{}\myworries{-based approaches} achieve the highest performance for all given time budgets\myworries{,  with only \ours{} w/ Hydra being more efficient for some of the small budgets, but not overall}. In contrast, random heuristics achieve very high coverage, resulting in good success rates, but often take very long to find specific objects. Further examining the different models, we find that \ours{} both travel much shorter distances and open fewer objects on average, indicating efficient and target-driven behavior. In contrast, Unstructured LLM produces almost 50\% more invalid actions. 
Qualitatively, we find that \ours{} is robust to various room layouts, such as "combined kitchen and living rooms" that result in large room clusterings and can handle the open-vocabulary room classification well. \myworries{In contrast, Hydra tends to predict a larger number of small rooms.} For reasoning examples, refer to \appref{app:analysis}.\looseness=-1

\vspace{-0.3em}
\subsection{Real-World Experiments}

We create a real-world apartment, consisting of four rooms: a combined kitchen and dining room, a living room, a long h
allway, and a bathroom. We use a Toyota HSR robot, equipped with an RGB-D camera and a \SI{270}{\degree} LiDAR. We replace the navigation policies with the ROS Nav Stack and the manipulation actions with the N$^2$M$^2$ manipulation policies~\cite{honerkamp2023learning}. We rely on the same assumptions as in simulation and assume access to localization, accurate semantic perception, and handle detection. We implement this by pre-recording a map with the robot's LiDAR and annotating it with semantic labels. At test time, we create an occupancy map from the robot's RGBD camera and reveal the corresponding part of the semantic map to the agent. The pre-recorded map is also used for localization. To detect handles, we use AR-Markers placed on each object. For details refer to the \appref{app:real_environment}.\looseness=-1 

We evaluate both \ours{} and the most efficient baseline, ESC\myworries{, on identical start positions and targets}. The results are shown in \tabref{tab:real_world_experiments}, \figref{fig:real_world}, and the video. Both methods succeeded in 8/10 episodes, demonstrating the successful transfer of the system to the real world. We find that the Voronoi- and scene graph construction transfer directly to the quite different, unseen layout. Similarly, the system directly transfers to the change in subpolicies, where the mobile manipulation policies ensure a the transition between all subpolicies.
The two failures stemmed from irrecoverable failures of the subpolicies, in particular, collisions of the base during navigation or of the arm while opening the door.
Comparing the methods, we find confirmation of the simulation results, with \ours{} moving and opening objects more target-driven and efficiently. Furthermore, the agent was able to react to the (unseen) subpolicy failures, such as re-trying to open a drawer when the gripper slipped off the handle.\looseness=-1

\input{tables/real_world_experiments}
\input{figures/tex-figures/real_world}

\vspace{-0.25cm}
\subsection{Towards General Household Tasks}
As we move to more abstract and complex tasks, it becomes increasingly difficult to define problem-specific rules or heuristics.
In contrast, our approach is readily expandable to a wide range of household and mobile manipulation tasks. Representative of this, we introduce a \textit{fuzzy search task}. In this task, the robot does not receive a specific object class to find, but rather a fuzzy description, such as  \textit{"I am hungry. Find me something for breakfast"}. The full set of queries are shown in \tabref{tab:fuzzy_search}.
We find that the agent is capable of finding objects that satisfy respective queries, and correctly reasoning about task completion by calling \textit{done()}.
We further test this capability with three tasks that cannot be solved with the given subpolicies (bottom part of \tabref{tab:fuzzy_search}). For these cases, the agent terminated the episode after finding the relevant objects, reasoning that these objects would now be sufficient for further completion of the tasks.
This demonstrates the flexibility of our approach. We leave the extension to arbitrary tasks to future work.\looseness=-1

%% file: tables/sim_experiments.tex
    \begin{table}
        \scriptsize
        \setlength{\tabcolsep}{4pt}
        \centering
        \caption{Interactive Object Search Results in Simulation}\label{tab:sim_experiments}
        \begin{threeparttable}
        \begin{tabularx}{\linewidth}{l|ccc|ccc}
          \toprule
            \multirow{2}{*}{Model} & \multirow{2}{*}{SR} & \multirow{2}{*}{SPL} & \multirow{2}{*}{AUC-E} & Object & Distance & Infeasible\\
             &  &  &  & Interactions & Traveled & Actions\\
          \midrule
          % \midrule
            Random           & 93.1 & 50.2 & 77.0 & 5.7 & 32.9 & --\\
            Greedy & 85.7 & 50.9 & 72.9 & 8.1 & 22.3 & -- \\
            \midrule
            ESC-Interactive & 95.4 & 62.7 & 84.5 & 4.1 & 19.6 & --\\
            HIMOS                  & 93.7 & 48.5 & 77.4 & 4.8 & 35.9 & --\\
            % SSG?                 && \\
            \midrule
            Unstructured LLM       & 86.3 & 59.4 & 77.6 & \underline{3.6} & 18.5 & 0.41 \\
            \rowfont{\myworriestable{}}
            \ours{} w/ Hydra & 92.0 & 61.9 & 84.3 & \textbf{2.7} & \textbf{12.9} & \textbf{0.06}\\
            \ours{} (ours)               & \textbf{97.7} & \textbf{63.6} & \textbf{87.2} & 3.9 & 18.2 & \underline{0.19} \\
            \midrule
            \midrule
            \rowfont{\myworriestable{}}
            Ours w/o frontiers  & 79.4 & 55.0 & 72.2 & 4.3 & 15.6 & 0.91 \\
            \rowfont{\myworriestable{}}
            Ours w/o history & 94.9 & \underline{63.0} & 84.1 & \underline{3.6} & 17.1 & 0.26 \\
            \rowfont{\myworriestable{}}
            Ours w/ room-history & \underline{97.1} & \underline{63.0} & \underline{86.6} & 3.8 & 17.8 & 0.28 \\
            \rowfont{\myworriestable{}}
            \rowfont{\myworriestable{}}
            Ours w/o distances & \underline{97.1} & 61.5 & 86.4 & 3.8 & 18.9 & 0.24 \\

          \bottomrule
        \end{tabularx}
        \vspace{-0.cm}
        \begin{tablenotes}[para,flushleft]
           \footnotesize      
           Top two in bold and underline. Object interactions, distance travelled and infeasible actions averaged over all episodes, including early terminated failures. Infeasible: avg. number of steps the LLM produced an action that could not be executed, resulting in re-planning with continued conversation (cf. Sec. IV-C.4).
         \end{tablenotes}
       \end{threeparttable}
       \vspace{-0.3cm}
    \end{table}

%% file: figures/tex-figures/auc_curve_pgf.tex
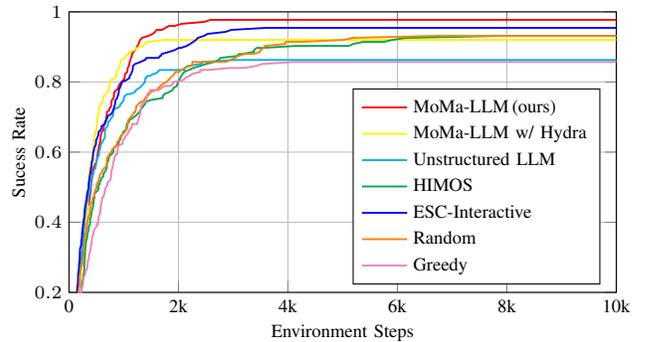
\begin{figure}[t]
\centering
\footnotesize
\begin{tikzpicture}
\begin{axis}[xlabel=Environment Steps,
ylabel=Sucess Rate, 
grid=both,
font=\scriptsize,
width=1.0\linewidth, % Adjust the width of the plot
height=0.6\linewidth, % Adjust the height of the plot
legend cell align=left,
legend pos=south east,
ymax=1, ymin=0.2,
xmin=0, xmax=10000,
xtick={0, 2000, 4000, 6000, 8000, 10000}, % Specify the x-tick positions
xticklabels={0, 2k, 4k, 6k, 8k, 10k}, % Specify the corresponding labels
scaled x ticks=false,
ylabel style={
      yshift=-2em, % Adjust the vertical distance for the y-label
    },
xlabel style={
      yshift=0.7em, % Adjust the vertical distance for the x-label
    },
label style={font=\scriptsize},
% axis on top,
% trim axis left,
% trim axis right,
% trim axis down,
% trim axis up,
% inner sep=0,
outer sep=0,
]
% ours
    \addplot[color=red, line width=0.7pt] coordinates {
(57,	0.005714285714286)
(71,	0.011428571428572)
(93,	0.017142857142857)
(99,	0.022857142857143)
(100,	0.028571428571429)
(104,	0.034285714285714)
(106,	0.04)
(108,	0.045714285714286)
(111,	0.051428571428572)
(115,	0.057142857142857)
(119,	0.062857142857143)
(120,	0.068571428571429)
(122,	0.074285714285714)
(122,	0.074285714285714)
(123,	0.08)
(126,	0.085714285714286)
(127,	0.091428571428572)
(130,	0.097142857142857)
(133,	0.102857142857143)
(135,	0.108571428571429)
(136,	0.114285714285714)
(137,	0.125714285714286)
(137,	0.125714285714286)
(139,	0.137142857142857)
(139,	0.137142857142857)
(140,	0.142857142857143)
(141,	0.148571428571429)
(144,	0.154285714285714)
(152,	0.16)
(155,	0.165714285714286)
(156,	0.182857142857143)
(156,	0.182857142857143)
(156,	0.182857142857143)
(157,	0.194285714285714)
(157,	0.194285714285714)
(159,	0.2)
(160,	0.205714285714286)
(169,	0.217142857142857)
(169,	0.217142857142857)
(173,	0.222857142857143)
(176,	0.228571428571429)
(179,	0.234285714285714)
(182,	0.24)
(183,	0.245714285714286)
(187,	0.251428571428571)
(194,	0.257142857142857)
(195,	0.262857142857143)
(201,	0.274285714285714)
(201,	0.274285714285714)
(205,	0.28)
(224,	0.285714285714286)
(229,	0.291428571428571)
(243,	0.297142857142857)
(254,	0.302857142857143)
(255,	0.308571428571429)
(256,	0.314285714285714)
(258,	0.32)
(260,	0.325714285714286)
(262,	0.331428571428571)
(266,	0.337142857142857)
(267,	0.342857142857143)
(268,	0.348571428571429)
(270,	0.365714285714286)
(270,	0.365714285714286)
(270,	0.365714285714286)
(273,	0.371428571428571)
(276,	0.377142857142857)
(279,	0.382857142857143)
(281,	0.388571428571429)
(282,	0.4)
(282,	0.4)
(292,	0.405714285714286)
(295,	0.411428571428571)
(296,	0.417142857142857)
(306,	0.422857142857143)
(313,	0.428571428571429)
(324,	0.434285714285714)
(327,	0.44)
(338,	0.445714285714286)
(361,	0.451428571428571)
(370,	0.457142857142857)
(371,	0.462857142857143)
(378,	0.468571428571429)
(380,	0.474285714285714)
(385,	0.48)
(393,	0.485714285714286)
(398,	0.491428571428571)
(399,	0.497142857142857)
(402,	0.502857142857143)
(405,	0.514285714285714)
(405,	0.514285714285714)
(415,	0.52)
(421,	0.525714285714286)
(423,	0.537142857142857)
(423,	0.537142857142857)
(428,	0.542857142857143)
(435,	0.548571428571429)
(445,	0.554285714285714)
(462,	0.56)
(480,	0.565714285714286)
(498,	0.571428571428571)
(509,	0.577142857142857)
(514,	0.582857142857143)
(518,	0.588571428571429)
(523,	0.594285714285714)
(531,	0.6)
(535,	0.605714285714286)
(544,	0.611428571428571)
(545,	0.617142857142857)
(569,	0.628571428571429)
(569,	0.628571428571429)
(572,	0.634285714285714)
(575,	0.64)
(579,	0.645714285714286)
(584,	0.651428571428571)
(600,	0.657142857142857)
(629,	0.662857142857143)
(636,	0.674285714285714)
(636,	0.674285714285714)
(640,	0.68)
(673,	0.685714285714286)
(677,	0.691428571428571)
(679,	0.697142857142857)
(687,	0.702857142857143)
(691,	0.708571428571428)
(692,	0.714285714285714)
(735,	0.72)
(743,	0.725714285714286)
(774,	0.731428571428571)
(775,	0.742857142857143)
(775,	0.742857142857143)
(782,	0.748571428571429)
(816,	0.754285714285714)
(818,	0.76)
(824,	0.765714285714286)
(825,	0.771428571428571)
(865,	0.777142857142857)
(901,	0.782857142857143)
(907,	0.788571428571429)
(928,	0.794285714285714)
(958,	0.8)
(993,	0.805714285714286)
(1017,	0.811428571428571)
(1018,	0.817142857142857)
(1024,	0.822857142857143)
(1066,	0.828571428571429)
(1068,	0.834285714285714)
(1072,	0.84)
(1093,	0.845714285714286)
(1101,	0.851428571428571)
(1109,	0.857142857142857)
(1120,	0.862857142857143)
(1131,	0.868571428571428)
(1149,	0.874285714285714)
(1150,	0.88)
(1178,	0.885714285714286)
(1212,	0.891428571428571)
(1234,	0.897142857142857)
(1249,	0.902857142857143)
(1273,	0.908571428571429)
(1282,	0.914285714285714)
(1300,	0.92)
(1333,	0.925714285714286)
(1449,	0.931428571428571)
(1528,	0.937142857142857)
(1571,	0.942857142857143)
(1600,	0.948571428571429)
(1700,	0.948571428571429)
(1761,	0.954285714285714)
(1825,	0.96)
(1937,	0.96)
(2051,	0.965714285714286)
(2474,	0.971428571428571)
(2585,	0.977142857142857)
(3771,	0.977142857142857)
(10000,	0.977142857142857)
        };

% hydra + ours
    \addplot[color=yellow, line width=0.7pt] coordinates {
(32,	0)
(32,	0)
(32,	0)
(32,	0)
(32,	0)
(32,	0)
(33,	0)
(33,	0)
(79,	0.005714285714286)
(89,	0.011428571428572)
(90,	0.017142857142857)
(97,	0.022857142857143)
(98,	0.034285714285714)
(98,	0.034285714285714)
(100,	0.04)
(101,	0.051428571428572)
(101,	0.051428571428572)
(102,	0.062857142857143)
(102,	0.062857142857143)
(103,	0.068571428571429)
(104,	0.074285714285714)
(105,	0.085714285714286)
(105,	0.085714285714286)
(107,	0.091428571428572)
(110,	0.097142857142857)
(111,	0.102857142857143)
(119,	0.108571428571429)
(121,	0.114285714285714)
(126,	0.12)
(129,	0.125714285714286)
(140,	0.131428571428571)
(143,	0.142857142857143)
(143,	0.142857142857143)
(146,	0.148571428571429)
(147,	0.154285714285714)
(152,	0.16)
(156,	0.165714285714286)
(161,	0.171428571428571)
(162,	0.177142857142857)
(164,	0.182857142857143)
(165,	0.188571428571429)
(176,	0.194285714285714)
(180,	0.2)
(187,	0.205714285714286)
(188,	0.211428571428571)
(193,	0.217142857142857)
(201,	0.222857142857143)
(204,	0.228571428571429)
(209,	0.234285714285714)
(213,	0.24)
(215,	0.245714285714286)
(218,	0.251428571428571)
(220,	0.257142857142857)
(226,	0.262857142857143)
(227,	0.268571428571429)
(240,	0.274285714285714)
(251,	0.28)
(252,	0.285714285714286)
(257,	0.291428571428571)
(266,	0.297142857142857)
(270,	0.302857142857143)
(273,	0.308571428571429)
(275,	0.314285714285714)
(276,	0.32)
(277,	0.331428571428571)
(277,	0.331428571428571)
(279,	0.337142857142857)
(283,	0.342857142857143)
(287,	0.348571428571429)
(297,	0.354285714285714)
(301,	0.36)
(302,	0.365714285714286)
(303,	0.371428571428571)
(304,	0.377142857142857)
(306,	0.382857142857143)
(307,	0.388571428571429)
(310,	0.394285714285714)
(312,	0.4)
(313,	0.405714285714286)
(315,	0.411428571428571)
(318,	0.417142857142857)
(324,	0.422857142857143)
(326,	0.428571428571429)
(329,	0.434285714285714)
(336,	0.445714285714286)
(336,	0.445714285714286)
(338,	0.445714285714286)
(344,	0.451428571428571)
(347,	0.468571428571429)
(347,	0.468571428571429)
(347,	0.468571428571429)
(348,	0.474285714285714)
(355,	0.485714285714286)
(355,	0.485714285714286)
(358,	0.491428571428571)
(359,	0.497142857142857)
(360,	0.502857142857143)
(361,	0.514285714285714)
(361,	0.514285714285714)
(364,	0.52)
(365,	0.525714285714286)
(366,	0.537142857142857)
(366,	0.537142857142857)
(373,	0.542857142857143)
(376,	0.548571428571429)
(377,	0.554285714285714)
(394,	0.56)
(403,	0.565714285714286)
(404,	0.577142857142857)
(404,	0.577142857142857)
(414,	0.582857142857143)
(426,	0.588571428571429)
(432,	0.594285714285714)
(436,	0.6)
(437,	0.605714285714286)
(438,	0.605714285714286)
(442,	0.611428571428571)
(449,	0.622857142857143)
(449,	0.622857142857143)
(457,	0.628571428571429)
(460,	0.634285714285714)
(462,	0.64)
(466,	0.645714285714286)
(483,	0.657142857142857)
(483,	0.657142857142857)
(484,	0.657142857142857)
(498,	0.662857142857143)
(499,	0.662857142857143)
(503,	0.668571428571429)
(504,	0.674285714285714)
(507,	0.68)
(513,	0.685714285714286)
(514,	0.691428571428571)
(519,	0.697142857142857)
(529,	0.702857142857143)
(531,	0.708571428571428)
(544,	0.714285714285714)
(549,	0.72)
(551,	0.725714285714286)
(563,	0.731428571428571)
(583,	0.737142857142857)
(616,	0.742857142857143)
(618,	0.748571428571429)
(620,	0.754285714285714)
(630,	0.76)
(675,	0.765714285714286)
(711,	0.771428571428571)
(722,	0.777142857142857)
(736,	0.788571428571429)
(736,	0.788571428571429)
(744,	0.794285714285714)
(749,	0.8)
(765,	0.805714285714286)
(837,	0.811428571428571)
(845,	0.811428571428571)
(859,	0.817142857142857)
(866,	0.822857142857143)
(868,	0.822857142857143)
(870,	0.828571428571429)
(889,	0.834285714285714)
(922,	0.84)
(928,	0.845714285714286)
(930,	0.851428571428571)
(945,	0.857142857142857)
(951,	0.862857142857143)
(1020,	0.868571428571428)
(1038,	0.874285714285714)
(1093,	0.88)
(1094,	0.885714285714286)
(1204,	0.891428571428571)
(1247,	0.897142857142857)
(1356,	0.902857142857143)
(1376,	0.908571428571429)
(1415,	0.914285714285714)
(1742,	0.92)
(10000,	0.92)
        };

% json-llm
  \addplot[color=cyan, line width=0.7pt] coordinates {
(33,	0)
(33,	0)
(33,	0)
(33,	0)
(57,	0.005714285714286)
(71,	0.011428571428572)
(91,	0.017142857142857)
(93,	0.022857142857143)
(97,	0.028571428571429)
(99,	0.034285714285714)
(100,	0.04)
(104,	0.045714285714286)
(106,	0.051428571428572)
(111,	0.057142857142857)
(112,	0.062857142857143)
(115,	0.068571428571429)
(118,	0.074285714285714)
(119,	0.08)
(122,	0.091428571428572)
(122,	0.091428571428572)
(123,	0.097142857142857)
(126,	0.102857142857143)
(127,	0.108571428571429)
(131,	0.114285714285714)
(133,	0.12)
(135,	0.125714285714286)
(136,	0.131428571428571)
(137,	0.137142857142857)
(145,	0.142857142857143)
(151,	0.154285714285714)
(151,	0.154285714285714)
(152,	0.16)
(154,	0.16)
(155,	0.165714285714286)
(156,	0.171428571428571)
(159,	0.177142857142857)
(160,	0.182857142857143)
(165,	0.188571428571429)
(169,	0.2)
(169,	0.2)
(172,	0.205714285714286)
(176,	0.211428571428571)
(179,	0.217142857142857)
(180,	0.222857142857143)
(182,	0.228571428571429)
(183,	0.234285714285714)
(187,	0.24)
(194,	0.245714285714286)
(195,	0.251428571428571)
(198,	0.257142857142857)
(199,	0.262857142857143)
(201,	0.274285714285714)
(201,	0.274285714285714)
(202,	0.274285714285714)
(206,	0.28)
(207,	0.285714285714286)
(217,	0.291428571428571)
(221,	0.297142857142857)
(228,	0.302857142857143)
(229,	0.308571428571429)
(230,	0.314285714285714)
(243,	0.32)
(247,	0.325714285714286)
(249,	0.331428571428571)
(249,	0.331428571428571)
(255,	0.337142857142857)
(256,	0.342857142857143)
(260,	0.348571428571429)
(262,	0.354285714285714)
(264,	0.365714285714286)
(264,	0.365714285714286)
(266,	0.371428571428571)
(267,	0.377142857142857)
(270,	0.388571428571429)
(270,	0.388571428571429)
(279,	0.394285714285714)
(281,	0.4)
(282,	0.411428571428571)
(282,	0.411428571428571)
(283,	0.417142857142857)
(287,	0.422857142857143)
(295,	0.428571428571429)
(305,	0.434285714285714)
(306,	0.44)
(313,	0.451428571428571)
(313,	0.451428571428571)
(324,	0.457142857142857)
(343,	0.462857142857143)
(347,	0.468571428571429)
(361,	0.474285714285714)
(371,	0.48)
(372,	0.485714285714286)
(378,	0.485714285714286)
(382,	0.485714285714286)
(383,	0.491428571428571)
(383,	0.491428571428571)
(386,	0.497142857142857)
(387,	0.502857142857143)
(391,	0.502857142857143)
(395,	0.508571428571429)
(399,	0.514285714285714)
(402,	0.52)
(405,	0.525714285714286)
(421,	0.525714285714286)
(423,	0.531428571428571)
(445,	0.537142857142857)
(446,	0.542857142857143)
(462,	0.548571428571429)
(468,	0.554285714285714)
(482,	0.554285714285714)
(485,	0.56)
(497,	0.565714285714286)
(497,	0.565714285714286)
(498,	0.571428571428571)
(507,	0.577142857142857)
(509,	0.582857142857143)
(540,	0.588571428571429)
(547,	0.594285714285714)
(549,	0.6)
(557,	0.605714285714286)
(574,	0.611428571428571)
(577,	0.617142857142857)
(583,	0.622857142857143)
(611,	0.628571428571429)
(626,	0.634285714285714)
(652,	0.64)
(655,	0.645714285714286)
(656,	0.651428571428571)
(659,	0.662857142857143)
(659,	0.662857142857143)
(674,	0.668571428571429)
(706,	0.674285714285714)
(712,	0.68)
(715,	0.685714285714286)
(752,	0.691428571428571)
(771,	0.697142857142857)
(802,	0.702857142857143)
(806,	0.708571428571428)
(811,	0.714285714285714)
(815,	0.72)
(858,	0.725714285714286)
(892,	0.725714285714286)
(931,	0.731428571428571)
(942,	0.737142857142857)
(979,	0.742857142857143)
(999,	0.748571428571429)
(1002,	0.754285714285714)
(1014,	0.76)
(1073,	0.765714285714286)
(1121,	0.765714285714286)
(1207,	0.771428571428571)
(1225,	0.777142857142857)
(1284,	0.782857142857143)
(1324,	0.788571428571429)
(1356,	0.794285714285714)
(1387,	0.8)
(1388,	0.805714285714286)
(1398,	0.811428571428571)
(1411,	0.811428571428571)
(1427,	0.817142857142857)
(1439,	0.817142857142857)
(1465,	0.817142857142857)
(1484,	0.817142857142857)
(1532,	0.822857142857143)
(1565,	0.822857142857143)
(1622,	0.828571428571429)
(1651,	0.834285714285714)
(2066,	0.834285714285714)
(2110,	0.834285714285714)
(2111,	0.84)
(2164,	0.845714285714286)
(2405,	0.851428571428571)
(2450,	0.857142857142857)
(3036,	0.862857142857143)
(4446,	0.862857142857143)
(10000,	0.862857142857143)
        };       
   
  % himos
  \addplot[color=Green, line width=0.7pt] coordinates {
        (108, 0.0114285714285714)
        (108, 0.0114285714285714)
        (118, 0.0171428571428571)
        (125, 0.0285714285714286)
        (125, 0.0285714285714286)
        (128, 0.0342857142857143)
        (130, 0.04)
        (132, 0.0457142857142857)
        (133, 0.0514285714285714)
        (139, 0.0571428571428571)
        (143, 0.0628571428571429)
        (146, 0.0685714285714286)
        (154, 0.08)
        (154, 0.08)
        (167, 0.0914285714285714)
        (167, 0.0914285714285714)
        (171, 0.0971428571428571)
        (178, 0.102857142857143)
        (186, 0.108571428571429)
        (189, 0.114285714285714)
        (192, 0.12)
        (194, 0.125714285714286)
        (196, 0.131428571428571)
        (199, 0.137142857142857)
        (200, 0.142857142857143)
        (201, 0.148571428571429)
        (202, 0.16)
        (202, 0.16)
        (209, 0.165714285714286)
        (214, 0.171428571428571)
        (220, 0.177142857142857)
        (223, 0.188571428571429)
        (223, 0.188571428571429)
        (224, 0.2)
        (224, 0.2)
        (227, 0.205714285714286)
        (229, 0.211428571428571)
        (232, 0.217142857142857)
        (238, 0.228571428571429)
        (238, 0.228571428571429)
        (249, 0.234285714285714)
        (256, 0.24)
        (258, 0.245714285714286)
        (265, 0.251428571428571)
        (274, 0.257142857142857)
        (276, 0.262857142857143)
        (278, 0.268571428571429)
        (283, 0.28)
        (283, 0.28)
        (285, 0.285714285714286)
        (287, 0.291428571428571)
        (288, 0.297142857142857)
        (291, 0.302857142857143)
        (293, 0.314285714285714)
        (293, 0.314285714285714)
        (294, 0.32)
        (300, 0.325714285714286)
        (301, 0.331428571428571)
        (311, 0.342857142857143)
        (311, 0.342857142857143)
        (314, 0.348571428571429)
        (330, 0.354285714285714)
        (338, 0.365714285714286)
        (338, 0.365714285714286)
        (342, 0.371428571428571)
        (356, 0.377142857142857)
        (357, 0.382857142857143)
        (361, 0.388571428571429)
        (384, 0.394285714285714)
        (385, 0.4)
        (386, 0.405714285714286)
        (389, 0.411428571428571)
        (404, 0.417142857142857)
        (407, 0.422857142857143)
        (413, 0.428571428571429)
        (418, 0.434285714285714)
        (429, 0.44)
        (440, 0.445714285714286)
        (443, 0.451428571428571)
        (445, 0.457142857142857)
        (451, 0.457142857142857)
        (452, 0.468571428571429)
        (452, 0.468571428571429)
        (462, 0.468571428571429)
        (464, 0.48)
        (464, 0.48)
        (491, 0.485714285714286)
        (530, 0.491428571428571)
        (535, 0.497142857142857)
        (537, 0.497142857142857)
        (540, 0.502857142857143)
        (545, 0.508571428571429)
        (558, 0.508571428571429)
        (560, 0.514285714285714)
        (584, 0.52)
        (592, 0.525714285714286)
        (628, 0.531428571428571)
        (629, 0.537142857142857)
        (647, 0.542857142857143)
        (675, 0.548571428571429)
        (688, 0.554285714285714)
        (694, 0.565714285714286)
        (694, 0.565714285714286)
        (703, 0.571428571428571)
        (720, 0.577142857142857)
        (725, 0.582857142857143)
        (733, 0.588571428571429)
        (783, 0.594285714285714)
        (807, 0.6)
        (813, 0.605714285714286)
        (814, 0.611428571428571)
        (861, 0.617142857142857)
        (902, 0.622857142857143)
        (908, 0.628571428571429)
        (929, 0.634285714285714)
        (944, 0.64)
        (945, 0.645714285714286)
        (980, 0.651428571428571)
        (986, 0.657142857142857)
        (1035, 0.662857142857143)
        (1047, 0.668571428571429)
        (1051, 0.674285714285714)
        (1054, 0.68)
        (1095, 0.685714285714286)
        (1097, 0.691428571428571)
        (1163, 0.697142857142857)
        (1210, 0.702857142857143)
        (1251, 0.708571428571428)
        (1286, 0.714285714285714)
        (1304, 0.72)
        (1331, 0.725714285714286)
        (1358, 0.731428571428571)
        (1389, 0.737142857142857)
        (1404, 0.742857142857143)
        (1489, 0.748571428571429)
        (1710, 0.754285714285714)
        (1717, 0.76)
        (1744, 0.765714285714286)
        (1833, 0.771428571428571)
        (1895, 0.777142857142857)
        (1909, 0.782857142857143)
        (1969, 0.788571428571429)
        (1980, 0.794285714285714)
        (1991, 0.8)
        (2013, 0.805714285714286)
        (2033, 0.811428571428571)
        (2052, 0.817142857142857)
        (2074, 0.822857142857143)
        (2105, 0.828571428571429)
        (2189, 0.834285714285714)
        (2275, 0.84)
        (2393, 0.845714285714286)
        (2506, 0.851428571428571)
        (2657, 0.857142857142857)
        (2735, 0.862857142857143)
        (2765, 0.868571428571429)
        (3066, 0.874285714285714)
        (3106, 0.88)
        (3169, 0.88)
        (3379, 0.885714285714286)
        (3383, 0.891428571428571)
        (3427, 0.897142857142857)
        (3542, 0.897142857142857)
        (4144, 0.902857142857143)
        (4585, 0.902857142857143)
        (4946, 0.902857142857143)
        (5114, 0.902857142857143)
        (5196, 0.908571428571429)
        (5375, 0.914285714285714)
        (5754, 0.914285714285714)
        (5908, 0.92)
        (6147, 0.925714285714286)
        (7922, 0.931428571428571)
        (8204, 0.931428571428571)
        (9716, 0.931428571428571)
        (10000, 0.931428571428571)
	};
% esc
  \addplot[color=blue, line width=0.7pt] coordinates {
(57,	0.005714285714286)
(73,	0.011428571428572)
(74,	0.017142857142857)
(78,	0.028571428571429)
(78,	0.028571428571429)
(86,	0.034285714285714)
(90,	0.04)
(91,	0.045714285714286)
(92,	0.051428571428572)
(93,	0.057142857142857)
(95,	0.062857142857143)
(96,	0.074285714285714)
(96,	0.074285714285714)
(97,	0.08)
(99,	0.091428571428572)
(99,	0.091428571428572)
(102,	0.097142857142857)
(106,	0.102857142857143)
(107,	0.114285714285714)
(107,	0.114285714285714)
(108,	0.125714285714286)
(108,	0.125714285714286)
(119,	0.131428571428571)
(121,	0.137142857142857)
(122,	0.142857142857143)
(126,	0.148571428571429)
(127,	0.154285714285714)
(129,	0.16)
(132,	0.165714285714286)
(133,	0.171428571428571)
(135,	0.177142857142857)
(137,	0.188571428571429)
(137,	0.188571428571429)
(139,	0.194285714285714)
(148,	0.2)
(149,	0.205714285714286)
(150,	0.211428571428571)
(154,	0.217142857142857)
(163,	0.228571428571429)
(163,	0.228571428571429)
(165,	0.245714285714286)
(165,	0.245714285714286)
(165,	0.245714285714286)
(166,	0.251428571428571)
(171,	0.257142857142857)
(180,	0.268571428571429)
(180,	0.268571428571429)
(182,	0.274285714285714)
(183,	0.28)
(189,	0.285714285714286)
(192,	0.291428571428571)
(194,	0.297142857142857)
(195,	0.302857142857143)
(198,	0.314285714285714)
(198,	0.314285714285714)
(199,	0.32)
(203,	0.325714285714286)
(216,	0.331428571428571)
(224,	0.337142857142857)
(225,	0.342857142857143)
(227,	0.348571428571429)
(233,	0.354285714285714)
(236,	0.36)
(239,	0.371428571428571)
(239,	0.371428571428571)
(247,	0.371428571428571)
(248,	0.382857142857143)
(248,	0.382857142857143)
(254,	0.388571428571429)
(256,	0.394285714285714)
(264,	0.4)
(265,	0.405714285714286)
(274,	0.411428571428571)
(281,	0.417142857142857)
(283,	0.417142857142857)
(284,	0.428571428571429)
(284,	0.428571428571429)
(285,	0.434285714285714)
(291,	0.44)
(295,	0.445714285714286)
(304,	0.451428571428571)
(315,	0.462857142857143)
(315,	0.462857142857143)
(323,	0.468571428571429)
(324,	0.474285714285714)
(330,	0.48)
(335,	0.48)
(337,	0.485714285714286)
(348,	0.491428571428571)
(356,	0.497142857142857)
(357,	0.502857142857143)
(361,	0.514285714285714)
(361,	0.514285714285714)
(370,	0.52)
(371,	0.525714285714286)
(380,	0.531428571428571)
(385,	0.537142857142857)
(390,	0.542857142857143)
(399,	0.548571428571429)
(402,	0.554285714285714)
(411,	0.56)
(413,	0.565714285714286)
(414,	0.571428571428571)
(433,	0.577142857142857)
(434,	0.582857142857143)
(437,	0.588571428571429)
(439,	0.594285714285714)
(443,	0.6)
(456,	0.605714285714286)
(459,	0.611428571428571)
(478,	0.617142857142857)
(491,	0.622857142857143)
(493,	0.628571428571429)
(496,	0.634285714285714)
(514,	0.64)
(529,	0.645714285714286)
(537,	0.651428571428571)
(540,	0.657142857142857)
(576,	0.662857142857143)
(594,	0.668571428571429)
(595,	0.674285714285714)
(655,	0.68)
(657,	0.685714285714286)
(696,	0.697142857142857)
(696,	0.697142857142857)
(707,	0.702857142857143)
(769,	0.708571428571428)
(794,	0.714285714285714)
(805,	0.72)
(821,	0.725714285714286)
(828,	0.731428571428571)
(840,	0.737142857142857)
(845,	0.742857142857143)
(850,	0.748571428571429)
(881,	0.754285714285714)
(889,	0.76)
(903,	0.765714285714286)
(909,	0.771428571428571)
(911,	0.777142857142857)
(926,	0.782857142857143)
(942,	0.788571428571429)
(960,	0.794285714285714)
(967,	0.8)
(1053,	0.805714285714286)
(1106,	0.811428571428571)
(1107,	0.817142857142857)
(1135,	0.822857142857143)
(1148,	0.828571428571429)
(1164,	0.834285714285714)
(1165,	0.84)
(1168,	0.845714285714286)
(1208,	0.851428571428571)
(1307,	0.857142857142857)
(1380,	0.862857142857143)
(1395,	0.862857142857143)
(1412,	0.868571428571428)
(1637,	0.868571428571428)
(1663,	0.874285714285714)
(1677,	0.874285714285714)
(1695,	0.88)
(1827,	0.885714285714286)
(1938,	0.891428571428571)
(2004,	0.897142857142857)
(2046,	0.897142857142857)
(2151,	0.902857142857143)
(2214,	0.908571428571429)
(2269,	0.914285714285714)
(2320,	0.92)
(2353,	0.925714285714286)
(2464,	0.931428571428571)
(2532,	0.937142857142857)
(2858,	0.942857142857143)
(2967,	0.948571428571429)
(3592,	0.954285714285714)
(4391,	0.954285714285714)
(10000,	0.954285714285714)
        };
   % random
  \addplot[color=orange, line width=0.7pt] coordinates {
(57,	0.005714285714286)
(91,	0.011428571428572)
(93,	0.017142857142857)
(96,	0.022857142857143)
(100,	0.028571428571429)
(102,	0.034285714285714)
(106,	0.04)
(107,	0.045714285714286)
(108,	0.051428571428572)
(110,	0.062857142857143)
(110,	0.062857142857143)
(113,	0.068571428571429)
(117,	0.074285714285714)
(122,	0.08)
(125,	0.085714285714286)
(126,	0.091428571428572)
(127,	0.097142857142857)
(128,	0.102857142857143)
(129,	0.108571428571429)
(133,	0.114285714285714)
(135,	0.12)
(137,	0.125714285714286)
(139,	0.137142857142857)
(139,	0.137142857142857)
(145,	0.142857142857143)
(147,	0.148571428571429)
(149,	0.154285714285714)
(154,	0.165714285714286)
(154,	0.165714285714286)
(157,	0.171428571428571)
(168,	0.177142857142857)
(169,	0.182857142857143)
(171,	0.2)
(171,	0.2)
(171,	0.2)
(177,	0.205714285714286)
(179,	0.211428571428571)
(185,	0.217142857142857)
(192,	0.222857142857143)
(193,	0.228571428571429)
(195,	0.234285714285714)
(196,	0.245714285714286)
(196,	0.245714285714286)
(204,	0.251428571428571)
(208,	0.257142857142857)
(209,	0.262857142857143)
(217,	0.268571428571429)
(224,	0.274285714285714)
(234,	0.28)
(238,	0.28)
(243,	0.285714285714286)
(255,	0.291428571428571)
(256,	0.297142857142857)
(262,	0.302857142857143)
(263,	0.308571428571429)
(266,	0.314285714285714)
(272,	0.32)
(273,	0.325714285714286)
(277,	0.331428571428571)
(279,	0.337142857142857)
(281,	0.342857142857143)
(282,	0.348571428571429)
(284,	0.354285714285714)
(294,	0.36)
(295,	0.365714285714286)
(310,	0.371428571428571)
(314,	0.377142857142857)
(320,	0.382857142857143)
(325,	0.388571428571429)
(327,	0.394285714285714)
(330,	0.394285714285714)
(338,	0.4)
(341,	0.405714285714286)
(353,	0.411428571428571)
(354,	0.417142857142857)
(377,	0.422857142857143)
(379,	0.428571428571429)
(383,	0.434285714285714)
(384,	0.44)
(392,	0.445714285714286)
(397,	0.451428571428571)
(420,	0.457142857142857)
(431,	0.462857142857143)
(433,	0.468571428571429)
(458,	0.474285714285714)
(464,	0.48)
(469,	0.485714285714286)
(476,	0.491428571428571)
(501,	0.497142857142857)
(503,	0.502857142857143)
(507,	0.508571428571429)
(525,	0.514285714285714)
(544,	0.52)
(551,	0.525714285714286)
(560,	0.531428571428571)
(569,	0.537142857142857)
(602,	0.542857142857143)
(609,	0.548571428571429)
(648,	0.554285714285714)
(680,	0.56)
(688,	0.565714285714286)
(692,	0.577142857142857)
(692,	0.577142857142857)
(693,	0.582857142857143)
(721,	0.588571428571429)
(734,	0.594285714285714)
(776,	0.6)
(781,	0.605714285714286)
(812,	0.611428571428571)
(857,	0.617142857142857)
(877,	0.622857142857143)
(878,	0.628571428571429)
(917,	0.634285714285714)
(925,	0.64)
(962,	0.645714285714286)
(978,	0.651428571428571)
(1021,	0.657142857142857)
(1039,	0.662857142857143)
(1064,	0.668571428571429)
(1066,	0.674285714285714)
(1072,	0.68)
(1073,	0.685714285714286)
(1083,	0.691428571428571)
(1112,	0.697142857142857)
(1143,	0.702857142857143)
(1171,	0.708571428571428)
(1190,	0.714285714285714)
(1207,	0.714285714285714)
(1219,	0.72)
(1225,	0.725714285714286)
(1229,	0.731428571428571)
(1268,	0.737142857142857)
(1307,	0.742857142857143)
(1325,	0.742857142857143)
(1377,	0.748571428571429)
(1384,	0.754285714285714)
(1434,	0.76)
(1500,	0.771428571428571)
(1500,	0.771428571428571)
(1579,	0.777142857142857)
(1626,	0.777142857142857)
(1630,	0.782857142857143)
(1647,	0.782857142857143)
(1666,	0.788571428571429)
(1677,	0.788571428571429)
(1705,	0.794285714285714)
(1783,	0.8)
(1850,	0.811428571428571)
(1850,	0.811428571428571)
(1891,	0.817142857142857)
(1921,	0.822857142857143)
(1943,	0.828571428571429)
(2013,	0.828571428571429)
(2051,	0.834285714285714)
(2116,	0.84)
(2205,	0.845714285714286)
(2250,	0.845714285714286)
(2252,	0.851428571428571)
(2255,	0.857142857142857)
(2867,	0.857142857142857)
(2946,	0.862857142857143)
(3028,	0.868571428571428)
(3094,	0.874285714285714)
(3206,	0.88)
(3377,	0.88)
(3492,	0.885714285714286)
(3496,	0.891428571428571)
(3565,	0.897142857142857)
(3575,	0.902857142857143)
(3952,	0.908571428571429)
(3957,	0.914285714285714)
(4243,	0.914285714285714)
(5022,	0.92)
(5176,	0.925714285714286)
(6687,	0.931428571428571)
(10000,	0.931428571428571)
        };
 
% greedy        
  \addplot[color=CarnationPink, line width=0.7pt] coordinates {
(57,	0.005714285714286)
(71,	0.011428571428572)
(91,	0.017142857142857)
(99,	0.022857142857143)
(100,	0.028571428571429)
(104,	0.034285714285714)
(106,	0.04)
(108,	0.045714285714286)
(113,	0.051428571428572)
(115,	0.057142857142857)
(118,	0.062857142857143)
(119,	0.068571428571429)
(122,	0.074285714285714)
(123,	0.08)
(126,	0.085714285714286)
(127,	0.091428571428572)
(129,	0.097142857142857)
(133,	0.102857142857143)
(135,	0.108571428571429)
(139,	0.114285714285714)
(147,	0.12)
(151,	0.125714285714286)
(155,	0.131428571428571)
(160,	0.137142857142857)
(165,	0.148571428571429)
(165,	0.148571428571429)
(169,	0.154285714285714)
(174,	0.16)
(182,	0.165714285714286)
(183,	0.171428571428571)
(187,	0.177142857142857)
(189,	0.177142857142857)
(201,	0.188571428571429)
(201,	0.188571428571429)
(217,	0.194285714285714)
(225,	0.2)
(227,	0.205714285714286)
(232,	0.211428571428571)
(243,	0.217142857142857)
(255,	0.222857142857143)
(256,	0.228571428571429)
(261,	0.234285714285714)
(267,	0.24)
(273,	0.245714285714286)
(292,	0.251428571428571)
(300,	0.257142857142857)
(308,	0.262857142857143)
(332,	0.268571428571429)
(335,	0.274285714285714)
(342,	0.28)
(345,	0.291428571428571)
(345,	0.291428571428571)
(362,	0.297142857142857)
(363,	0.302857142857143)
(367,	0.308571428571429)
(382,	0.314285714285714)
(383,	0.314285714285714)
(395,	0.325714285714286)
(395,	0.325714285714286)
(400,	0.331428571428571)
(414,	0.342857142857143)
(414,	0.342857142857143)
(418,	0.348571428571429)
(419,	0.354285714285714)
(425,	0.36)
(446,	0.365714285714286)
(447,	0.371428571428571)
(457,	0.377142857142857)
(475,	0.382857142857143)
(509,	0.388571428571429)
(510,	0.394285714285714)
(514,	0.394285714285714)
(523,	0.394285714285714)
(531,	0.4)
(532,	0.405714285714286)
(537,	0.411428571428571)
(545,	0.417142857142857)
(564,	0.422857142857143)
(570,	0.428571428571429)
(574,	0.434285714285714)
(578,	0.44)
(581,	0.44)
(582,	0.445714285714286)
(588,	0.451428571428571)
(596,	0.457142857142857)
(600,	0.462857142857143)
(602,	0.468571428571429)
(606,	0.468571428571429)
(629,	0.474285714285714)
(629,	0.474285714285714)
(642,	0.48)
(649,	0.485714285714286)
(660,	0.491428571428571)
(666,	0.497142857142857)
(687,	0.502857142857143)
(702,	0.508571428571429)
(707,	0.514285714285714)
(753,	0.52)
(764,	0.525714285714286)
(769,	0.531428571428571)
(771,	0.542857142857143)
(771,	0.542857142857143)
(779,	0.548571428571429)
(792,	0.554285714285714)
(799,	0.56)
(805,	0.565714285714286)
(807,	0.571428571428571)
(827,	0.577142857142857)
(829,	0.582857142857143)
(840,	0.588571428571429)
(854,	0.594285714285714)
(872,	0.594285714285714)
(877,	0.594285714285714)
(886,	0.6)
(891,	0.611428571428571)
(891,	0.611428571428571)
(897,	0.617142857142857)
(955,	0.622857142857143)
(967,	0.622857142857143)
(983,	0.628571428571429)
(985,	0.634285714285714)
(1020,	0.64)
(1033,	0.645714285714286)
(1069,	0.651428571428571)
(1084,	0.657142857142857)
(1103,	0.662857142857143)
(1108,	0.662857142857143)
(1133,	0.668571428571429)
(1136,	0.668571428571429)
(1149,	0.674285714285714)
(1153,	0.68)
(1158,	0.68)
(1245,	0.68)
(1251,	0.685714285714286)
(1274,	0.697142857142857)
(1274,	0.697142857142857)
(1288,	0.702857142857143)
(1298,	0.708571428571428)
(1302,	0.714285714285714)
(1311,	0.72)
(1316,	0.725714285714286)
(1317,	0.731428571428571)
(1337,	0.737142857142857)
(1390,	0.742857142857143)
(1407,	0.748571428571429)
(1420,	0.754285714285714)
(1421,	0.76)
(1430,	0.765714285714286)
(1478,	0.771428571428571)
(1483,	0.777142857142857)
(1493,	0.777142857142857)
(1639,	0.777142857142857)
(1693,	0.782857142857143)
(1738,	0.788571428571429)
(1742,	0.788571428571429)
(1825,	0.788571428571429)
(1845,	0.788571428571429)
(1896,	0.794285714285714)
(1900,	0.8)
(1917,	0.8)
(2092,	0.805714285714286)
(2119,	0.811428571428571)
(2192,	0.817142857142857)
(2206,	0.817142857142857)
(2246,	0.822857142857143)
(2396,	0.828571428571429)
(2399,	0.834285714285714)
(2494,	0.834285714285714)
(2554,	0.834285714285714)
(2932,	0.84)
(3210,	0.84)
(3410,	0.845714285714286)
(3528,	0.851428571428571)
(4120,	0.857142857142857)
(6575,	0.857142857142857)
(10000,	0.857142857142857)
        };
 \legend{\ours\,(ours), \ours{} w/ Hydra, Unstructured LLM, HIMOS, ESC-Interactive, Random, Greedy};
\end{axis}
\end{tikzpicture}
\vspace{-0.3cm}
\caption{Interactive search efficiency curve in simulation. Each point depicts the success rate for a given maximum time budget (x-axis).}
\label{fig:auc-vis-pgf}
\vspace{-0.4cm}
\end{figure}

%% file: tables/env_partitioning.tex
    \begin{table}
    \vspace{-0.2cm}
    \setlength{\tabcolsep}{2pt}
        \centering
        \caption{Environment Partitioning Throughout Exploration}
        \vspace{-0.2cm}
        \myworriestable{}
        \label{tab:igibson_env_partitioning}%
        \begin{threeparttable}
        \begin{tabularx}{\linewidth}{l|YYYYYY}
          \toprule
            \multirow{2}{*}{Approach} & \multicolumn{2}{c}{Precision}  & \multicolumn{2}{c}{Recall} & \multirow{2}{*}{Purity $\uparrow$} \\ % & Room \\
            & $\mu\uparrow$ & $\sigma\downarrow$ & $\mu\uparrow$ & $\sigma\downarrow$ & \\ % & Semantics \\
          \midrule
            Hydra & 0.621 & 0.081 & 0.943 & 0.044 & 0.562 \\ % & - \\
            \ours{}  & \textbf{0.666} & \textbf{0.064} & \textbf{0.948} & \textbf{0.032} & \textbf{0.615} \\ % & 0.276 \\
          \bottomrule
        \end{tabularx}
        \begin{tablenotes}[para,flushleft]
           \footnotesize      
           Dense room segmentation precision and recall as defined in Hughes~\textit{et~al.}~\cite{hughes2022hydra} in terms of mean and standard deviation throughout exploration. The purity (Supplementary Sec. S.6.1.A) measures the number of ground-truth rooms erroneously captured per predicted room given sparse Voronoi graphs. Evaluated across 10 episodes and all test scenes with 2D grid resolution of \SI{0.05}{m} to account for thin walls. Best values are written in bold.
         \end{tablenotes}
       \end{threeparttable}
    \vspace{-0.3cm}
    \end{table}

%% file: tables/real_world_experiments.tex
    \begin{table}
    \vspace{-0.2cm}
        \centering
        \caption{Interactive Object Search Results in the Real World }
        \vspace{-0.2cm}
        \label{tab:real_world_experiments}%
        \begin{threeparttable}
        \begin{tabularx}{\linewidth}{l|YYYYY}
          \toprule
            \multirow{2}{*}{Model} & Success & Navig & Manip & Distance & Object \\
                                   & Rate & Fails & Fails & Traveled & Interact. \\
          \midrule
            ESC-Inter. & 80\% & 2 & 0 & 33.9 & 3.5 \\
            \ours{}    & 80\% & 1 & 1 & \textbf{17.9} & \textbf{2.2} \\
          \bottomrule
        \end{tabularx}
        \begin{tablenotes}[para,flushleft]
           \footnotesize      
           Dist. travelled is the average distance travelled per episode in meters. Object interactions are the average number of object interactions per episode.
         \end{tablenotes}
       \end{threeparttable}
    \vspace{-0.25cm}
    \end{table}

%% file: figures/tex-figures/real_world.tex
\setlength{\tabcolsep}{1pt}
\begin{figure}[t]
	\centering
	\resizebox{0.7\linewidth}{!}{%
 \includegraphics[width=\linewidth,trim={2cm 17cm 2cm 17cm},clip,angle =0,valign=c]{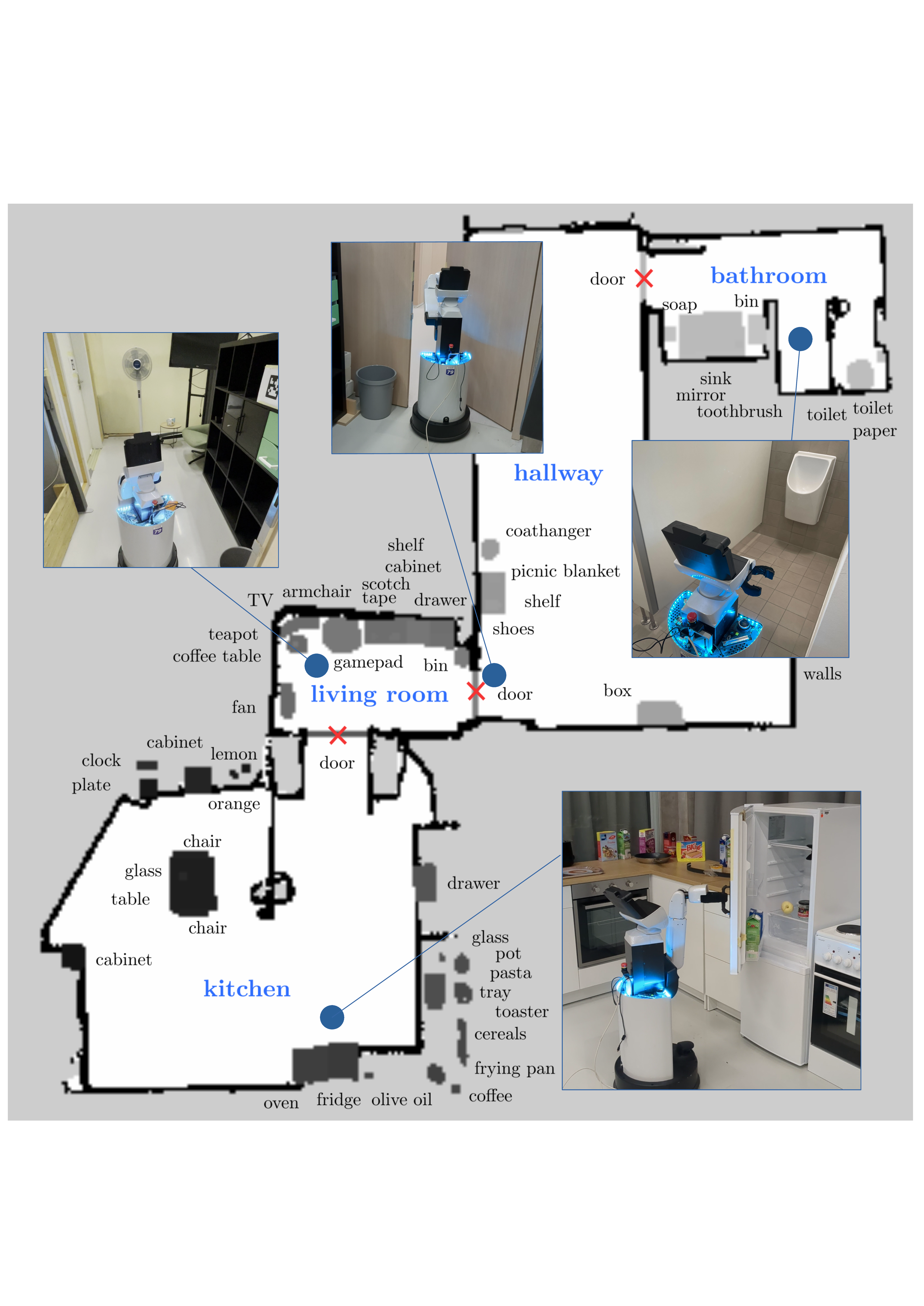}
 }
	% \vspace{-0.2cm}
     \caption{We construct a real-world apartment covering four rooms and 54 objects and transfer the model to a Toyota HSR robot.} 
  	\label{fig:real_world}
   \vspace{-0.4cm}
\end{figure}
\setlength{\tabcolsep}{6pt}

%% file: sections/5_conclusion.tex
\vspace{-0.2cm}
\section{Conclusion}
We developed a method to ground language models for high-level reasoning with scalable, dynamic scene graphs and efficient low-level policies for interactive tasks that require combined reasoning about manipulation, navigation and exploration. We demonstrated the importance of extracting structured knowledge for large and unexplored scenes to enable LLMs to reason about efficient search strategies, outperforming fully learned or co-occurrence-based methods. We then transfer\myworries{r}ed our method to a real-world apartment, achieving consistent performance over many episodes.
Lastly, we demonstrate the extendability of our approach to abstract tasks, opening the door towards general household tasks. 

%% file: sections/6_appendix.tex
%%%%%%%%%% Merge with supplemental materials %%%%%%%%%%

\clearpage
\renewcommand{\baselinestretch}{1}
\setlength{\belowcaptionskip}{0pt}

\begin{strip}
\begin{center}
\vspace{-5ex}
\textbf{\LARGE \bf
% Grounding Language with Dynamic Scene Graphs \\\vspace{0.5ex}for Interactive Object Search with Mobile Manipulators
Language-Grounded Dynamic Scene Graphs for \\\vspace{0.5ex} Interactive Object Search with Mobile Manipulation
} \\
\vspace{3ex}

\Large{\bf- Supplementary Material -}\\
\vspace{0.4cm}
\normalsize{Daniel Honerkamp$^{1*}$, Martin Büchner$^{1*}$, Fabien Despinoy$^{2}$, Tim Welschehold$^{1}$, Abhinav Valada$^{1}$}\\
\end{center}
\end{strip}

%%%%%%%%%% Merge with supplemental materials %%%%%%%%%%
\setcounter{section}{0}
\setcounter{equation}{0}
\setcounter{figure}{0}
\setcounter{table}{0}
\setcounter{page}{1}
\makeatletter

\renewcommand{\thesection}{S.\arabic{section}}
\renewcommand{\thesubsection}{S.\arabic{subsection}}
\renewcommand{\thetable}{S.\arabic{table}}
\renewcommand{\thefigure}{S.\arabic{figure}}

\let\thefootnote\relax\footnote{$^*$These authors contributed equally.\\$^1$Department of Computer Science, University of Freiburg, Germany.\\$^2$Toyota Motor Europe (TME)\\
Project page: \website{}
}%
\normalsize 
In this supplementary material, we provide additional details on the simulation and real-world environments, the subpolicies, and the baselines. Moreover, we provide additional results from the experiments. Further examples of real-world experiments are also demonstrated in the video material.

\section{Simulation Environment}\label{app:environment}
\subsection{Modifications}
We make the following modifications to the iGibson scenes:
\begin{itemize}[topsep=0pt]
    \item We close all exterior doors and filter them out of the scene graph as they lead out to empty space.
    \item We scale the size of the Fetch robot by a factor of 0.85 as otherwise it is too large to navigate a significant number of areas.
    \item We remove six doors that block the robot's pathway when opening (\textit{door\_52} in \textit{Pomaria\_0\_int}, \textit{door\_75} in scene \textit{Beechwood\_1\_int} and \textit{door\_106} in \textit{Ihlen\_1\_int}, \textit{door\_90} in \textit{Merom\_1\_int}, \textit{door\_138} and \textit{door\_139} in \textit{Wainscott\_0\_int}).
    \item We rename four object categories, with misleading asset names, such as renaming \textit{breakfast\_table} (which can be found e.g. in bedrooms) to  \textit{table}.
\end{itemize}

\subsection{Task Generation}
We first align the rooms and objects with the prior distribution. We manually match room labels and use cosine similarities of object name embeddings to match objects. We embed the object names from the scene and from the prior distribution with SBERT~\citeS{reimers2019sentence}. We then define a match as a cosine similarity $\le 0.7$ and being in the top 50 matched categories.
As $P^{prior}$ only contains two "inside" relationships, we enrich these relationships by assuming that all objects that can be found on top of an object and that fit in size, can also be found inside it and vice-versa. This results in an extended distribution $P^{prior, ext}$.
We then procedurally enrich each episode by drawing additional objects from the set of admissible room-object-relations for all existing furniture in the iGibson scenes. We keep drawing until the physical placement of a relation succeeds or the distribution is exhausted.
Given a valid scene instantiation, we draw a target category $g \sim U(scene)$ from all categories in the scene. We reject targets as infeasible if no target instance is reachable from the agent's random start position. We also reject all episodes in which the target object is immediately visible.

\subsection{Perception}
The robot in simulation is equipped with a differential drive and an RGB-D camera with a resolution of 256 $\times$ 256 pixels and a vertical field of view of $\SI{120}{\degree}$. As the focus of this work is on decision making, we abstract from imperfect perception and assume access to ground truth instances and semantic segmentation from the simulator. For a realistic detection range, we restrict all sensors (depth, semantics) to a maximum range of $\SI{5}{\meter}$ and set a threshold of 50 segmented pixels before an object is considered detected - except for objects with a volume below \SI{0.01}{\cubic\metre}.
We furthermore assume accurate detection of whether an articulated object is open or closed and assume that objects within receptacles are detected after opening the object.
We construct all maps at a resolution of \SI{0.075}{\meter} and detect the \textit{floor} and \textit{carpet} categories as free space.

\subsection{Execution}
All methods start by fully turning around in-place to initialize the scene graph.

\para{Low-level action space:}
The subpolicies act in a low-level action space consisting of the following actions:
\begin{itemize}[topsep=0pt]
    \item move forward by \SI{7.5}{\cm}
    \item turn-left by up to \SI{35}{\radian}
    \item turn-right by up to \SI{35}{\radian}
    \item open articulated object
    \item close articulated object
    \item done: end the episode and evaluate the success
\end{itemize}

\para{Navigation:} To reduce run-time, navigation actions are executed without physical simulation, but rather by directly setting the state of the robot sequentially to each waypoint the resulting path from the A$^*$-planner (matching the map resolution of \SI{0.075}{\meter}), collecting the observations along the whole path. If no complete path in free space exists, the navigation policy will consider unexplored areas as traversable and replan with newly revealed free space. It will return 'failed' if it cannot find a path or reaches too many replanning attempts.

\para{Object interactions}:
We follow previous work~\citeS{shridhar2020alfred, schmalstieg2023learning} and execute articulated object manipulations as "magic actions". These actions directly set the joint values of the object to their minimum or maximum. In the calculation of the efficiency curves and AUC-E, we weigh these actions by a time factor of 30 to make their time cost comparable to an execution duration of roughly 30 seconds.

The $go\_to\_and\_open()$ action is implemented as follows:
\begin{enumerate}[topsep=0pt]
    \item Navigate to the Voronoi node closest to the object
    \item Navigate to the most central free space in front of the object and turn toward it
    \item Trigger the magic open action
\end{enumerate}

\section{Real-World Environment}\label{app:real_environment}

\subsection{Map}
\figref{fig:real-world-map} shows the map of the real-world environment. Different object instances are indicated in different intensities. The environment covers five different room types and 54 different object categories, including furniture such as tables, chairs, and a coat hanger, as well as small and less common objects such as soap, gamepad, or scotch tape.
We set start positions for the robot in the kitchen, living room, and hallway. We ensure the same initial positions and targets for all methods. The target categories given to the agents were \textit{shoes}, \textit{milk} (inside fridge), \textit{knife} (inside kitchen drawer), \textit{book} (inside living room cabinet), \textit{toilet paper}, \textit{toaster}, \textit{tea} (inside kitchen cabinet), \textit{pencil} (inside living room drawer), \textit{soap} and \textit{lemon}.

\input{figures/tex-figures/real-world-map}

\subsection{Execution}
\para{Navigation:} We use the default manufacturer versions of the ROS NavStack as developed by Toyota for navigation. It uses the robot's LiDAR and depth cameras for dynamic obstacle detection and navigates in a map inflated by \SI{0.25}{\meter}.

\para{Mobile manipulation:}
Articulated object interactions are executed with pretrained N$^2$M$^2$ manipulation policies~\citeS{honerkamp2023learning}. The policy receives the handle pose, detected through AR-Markers, and uses a local occupancy map constructed from the LiDAR of the robot for obstacle avoidance. For each articulated object, we collect a single demonstration of opening the object. This demonstration consists of a set of poses of the robot's wrist link during the opening motion. These poses are then interpolated with splines to generate an end-effector motion. This agent's aim is to follow this end-effector motion to complete the object interaction.
We evaluate whether the motion was successful based on whether the marker on the object changed its position after the execution of the subpolicy.
For doors, we do not lock the spring-loaded door latch, as the robot is not strong enough to press the handle down.

The only exception to this is the door between the kitchen and living room. We found that the robot was unable to localize itself during the opening motion, as the moving door occupies the overwhelming majority of the robot's LiDAR measurements. As a result, it was not possible to follow the opening demonstration whenever the marker moved out of the robot's view. Instead, we use a simpler pushing motion from the inside and abstain from opening it in the other direction.

\section{Hierachical Scene Graph Structure}
\input{tables/scene_graph_structure}
In order to provide a concise overview a complete picture of the employed scene graph hierarchy we provide a detailed overview of it in \tabref{tab:scene_graph_structure}.

In addition, we outline the construction of the navigational Voronoi graph used for navigation and associating objects to regions in the following. The Generalized Voronoi Diagram (GVD)~\citeS{choset1995sensor} comprises two-equidistant faces that each represent the set of points equidistant to two obstacles $C_i$ and $C_j$. Each point in this set is closer to $C_i$ and $C_j$ than any other obstacle:
\begin{equation}
    \mathcal{F}_{ij} = \{ x \in \mathbb{R}^{m} : 0 \leq d_{i}(x) = d_{j}(x) \forall k \neq i,j, \nabla d_{i}(x) \neq d_{j}(x) \}.
\end{equation}
The union of all two-equidistant faces generated by the obstacle positions defined by $\mathcal{B}_{t}$ is called the 2-Voronoi set $\mathcal{F}^{2}$ or the two-dimensional GVD of the space of obstacles $C_i \in \mathcal{B}_{t}$:
\begin{equation}
    \mathcal{F}^{2} = \overset{n-1}{ \underset{i=1}{\cup}} 
    \overset{n}{ \underset{j=i+1}{\cup}} \mathcal{F}_{ij}.
\end{equation}
The set of points contained in $\mathcal{F}^{2}$ constitutes the initial set of edges of the generalized Voronoi graph (GVG). We extract the corresponding nodes by computing the 3-Voronoi sets, which constitute the joints of the GVD:
\begin{equation}
    \mathcal{F}^{3} = \overset{n-2}{ \underset{i=1}{\cup}}
    \overset{n-1}{ \underset{j=i+1}{\cup}} 
    \overset{n}{ \underset{k=j+1}{\cup}} \mathcal{F}_{ijk}.
\end{equation}
Given this, the generalized Voronoi graph GVG = \{$\mathcal{F}^{2}, \mathcal{F}^{3}$\} undergoes sparsification by eliminating edges of degree 2 to form $\mathcal{G}_{\mathcal{V}}$.

\input{tables/distance_encoding}
\section{Language encoding}\label{app:language}
We encode distance to natural language based on a discrete mapping, following the principle of~\citeS{chalvatzaki2023learning}. We bin the distance to the object, then apply the mapping reported in \tabref{tab:distance_encoding}. This results in a consistent relative language encoding.

\input{figures/tex-figures/json-policy}
\section{Baselines}\label{app:baselines}
The \textit{Unstructured LLM} baseline receives the same instructions and "remember" notes as our approach. The full JSON-formatted prompt of this baseline is depicted in \figref{fig:json-policy}. We find that the much less structured and longer prompt leads to more frequent invalid actions or hallucinations (cf.~\tabref{tab:sim_experiments}), such as trying to open objects that do not exist or are already opened. If stuck for repeated steps, this can result in failed episodes. A second source of failures are wrong terminations, in which the LLM calls \textit{done()} while it has not found the correct object of interest.

\section{Additional Results}\label{app:analysis}

\subsection{Hierarchical Scene Graph}
\label{app:scene_graph_acc}

\subsubsection{Metrics}
\label{sec:scene_graph_metrics}
In the following, we list the metrics used for evaluating the accuracy of the scene graph.

\noindent\textit{Room Segmentation Precision / Recall:} In order to compare our method of room segmentation against the approach used by Hydra~\citeS{hughes2022hydra} we make use of the same metrics they evaluated:
\begin{equation}
    P_t =  \frac{1}{|R_{e}|} \sum_{r_e \in R_e} \underset{{r_g \in R_g}}{\operatorname{max}} \frac{|r_g \cap r_e|}{|r_e|},
\end{equation}
\begin{equation}
    R_t = \frac{1}{|R_{g}|} \sum_{r_g \in R_g} \underset{{r_e \in R_e}}{\operatorname{max}} \frac{|r_e \cap r_g|}{|r_g|}, 
\end{equation}
where $R_e$ is the set of estimated rooms and $R_g$ is the set of
ground-truth rooms. The cardinality of a set is given by $|\cdot|$. Each room $r_e$ or $r_g$ is defined by its set of covered pixels on a 2D grid. In order to reflect instabilities throughout exploring the environment we report the means and standard deviations of the precision and recall, respectively:
\begin{equation}
    \bar{P} = \frac{1}{T} \sum_{t=1}^{T} P_t  \quad\quad \sigma_P = \sqrt{\frac{\sum_{t=1}^{T} (P_t - \bar{P})^{2}}{T}}
\end{equation}
\begin{equation}
    \bar{R} = \frac{1}{T} \sum_{t=1}^{T} R_t \quad\quad \sigma_R = \sqrt{\frac{\sum_{t=1}^{T} (R_t - \bar{R})^{2}}{T}}
\end{equation}
The metrics are evaluated on a dense 2D grid with a resolution of \SI{0.05}{m}, which is increased compared to the normal resolution of \SI{0.075}{m} used in all other evaluations. This is done to account for thin walls contained in iGibson. In order to evaluate the separated Voronoi graphs covering distinct rooms on a dense manifold we employ room-specific wavefronts initialized at each node of the separated Voronoi graph bounded by the extracted obstacles.

\noindent\textit{Room Segmentation Purity}: In addition to the dense evaluation outlined above we also evaluate the purity of each of the generated components of the separated Voronoi graphs. Being a criterion used for measuring clustering quality~\citeS{manning2008introduction} it penalizes the effect of covering multiple ground truth rooms per classified room.
\begin{equation}
    purity(\Omega, \mathbb{C}) = \frac{1}{N} \sum_{k} \operatorname{max}_{j} | \omega_k \cap c_j |
\end{equation}
where $\Omega = \{\omega_1, \omega_2, \ldots, \omega_{K}\}$ is the set of components of $\mathcal{G}_{V}^{R}$ and $\mathbb{C}=\{c_1, c_2, \ldots, c_{J}\}$ is the set of ground-truth rooms. Each $\omega_k$ holds the predicted room types of all nodes of the respective component. In our case, the purity measures the extent to which a set of Voronoi nodes covers a room that contains a single class. Thus, the graph purity describes the degree of room under-segmentation apparent in the scene and thus measures how well the door-wise Voronoi graph separation performs. Similar to the room segmentation precision and recall we average the purity throughout exploring the environment.

\input{tables/closed_set_room_eval}

\subsubsection{Room Segmentation and Classification}
Our proposed room separation scheme relies on separating Voronoi graphs at door positions. Thus, it is prone to under-segmentation whenever faced with open room layouts or, e.g., \textit{missing} doors to hallways. We visualize the scene graphs produced by Hydra as well as \ours{} in \figref{fig:hydra-vs-momallm-sg}. Regarding the approach of Hydra, we observe that maps containing a many corridors with similar diameters produce either a very large number or a very small number of graph components when applying a range of various obstacle dilation values. As Hydra selects the final environment partitioning based on the median of the number of graph components obtained through various obstacle dilation values, it is faced with a bi-modal distribution. This ultimately renders a median-based selection of the segmentation difficult.

\input{figures/tex-figures/hydra-vs-momallm-sg}

In addition to evaluation in \tabref{tab:igibson_env_partitioning}, we observe an average purity for \ours{} of 0.615 throughout the exploration over 10 episodes across all of the iGibson test scenes. Compared with that, the room segmentation approach introduced by Hydra~\citeS{hughes2022hydra} reaches a purity of 0.562 as listed in \tabref{tab:igibson_env_partitioning}. While both Hydra and \ours{} tend to under-segment the given room layout, Hydra is specifically affected by narrow constrictions induced by obstacle placements and non-varying door widths, which creates a significant number of isolated graphs covering small corridors that do not represent full ground-truth regions. In general, we infer that long and narrow corridors as well as cluttered scenes are challenging to segment using classical morphological segmentation algorithms (see \figref{fig:hydra-vs-momallm-sg}).

We found our downstream policy to be robust to under-segmented rooms even though objects from multiple rooms were, e.g., considered part of a single room. By relying on the camera pose from which an object is observed we reduce the number of false object-room assignments (\textit{through} walls) to a minimum. We show multiple resulting Voronoi graphs in \figref{fig:examples}.

\input{figures/tex-figures/examples}

In addition to the time-wise averaged room segmentation results reported in \tabref{tab:igibson_env_partitioning}, we evaluate the semantic room categories predicted by GPT-3.5. in \tabref{tab:room_cat_eval}. Even though \ours{} normally uses open-set room categories, we evaluate the performance on a closed-set of room categories to report reproducible results. To do so, we provided GPT-3.5 with all room categories contained in the iGibson dataset with the task to pick the most suitable given the objects assigned to each particular Voronoi component representing a room. Similar to the segmentation evaluation, we report numbers that are averaged over 10 episodes per scene as well as across all high-level policy steps per episode. We compare the predicted room category of each Voronoi node with the underlying ground-truth room layout maps.
Following this, we arrive at an average predicted room category accuracy of 27.6\% This number is largely affected by open room layouts as mentioned above. 

In addition to the closed-set evaluation, we also evaluated the predicted room categories in an open-set manner on the real-world map shown in \figref{fig:real-world-map}. Across the 10 trials executed in the real world as given in \tabref{tab:real_world_experiments}, we follow the same evaluation protocol and obtain an average room category accuracy of 90.1\% as listed in~\tabref{tab:room_cat_eval}. Human-level assessment allows evaluating errors such as \textit{entryway} instead of \textit{hallway} positively, which drastically increases the metrics. Nonetheless, the real-world map is less complex in terms of its topology and object distribution compared to the iGibson environments, which feature, e.g., rooms with no objects contained.

\subsection{Reasoning} 
\tabref{tab:fuzzy_search} show the full set of fuzzy search queries (top) and infeasible queries (bottom) that were evaluated in \secref{sec:experiments}, together with the language model's reasoning in response to these tasks. 

\figref{fig:examples} shows additional examples of the scene representations and the model reasoning. It depicts the Voronoi graph and frontiers to unexplored areas (left), the BEV-map together with the constructed scene graph (middle), and the input prompt and answers of the LLM (right). Additional video material with full prompt reasoning is shown on the project website.

\input{tables/fuzzy_search}

\subsection{Deployment with Full Perception Pipeline}
While we focus on evaluation with ground-truth perception in the main work to be able to focus on the representation and decision making components, this section provides details on the requirements of full deployment and guidance to facilitate the deployment.

Our approach requires (i) an RGB-D sensor (ii) localization and mapping (iii) semantic segmentation and (iv) grasp pose detection. Modern RGB-D SLAM approaches such as RTAB-Map~\citeS{labbe2019rtab} can provide (i) and (ii).
As our approach supports open-vocabulary representation and reasoning, it enables deployment with any semantic segmentation model, irrespective of its supported classes. This can be closed vocabulary methods~\citeS{vodisch2023codeps, gosala2023skyeye} such as Mask R-CNN~\citeS{he2017mask} or newer, transformer-based methods~\citeS{cheng2021mask2former, kappeler2023few}. A further range of methods can provide object detections and bounding boxes for a given list of open-vocabulary query categories~\citeS{gu2022openvocabulary, minderer2022simple, minderer2024scaling, kuo2023openvocabulary}. The best model should be based on the use-case, available compute, required object categories, and accuracy.
Chen et al.~\cite{chen2023not} provide a possible reference implementation: they deploy an RTAB-node for localization and mapping. This can be extended to semantic labels through an additional RTAB-node that listens to the semantic masks, to then fuse the resulting point clouds.
Finally (iv), for handle detection and grasp-pose detection, \citeS{arduengo2021robust} achieve accurate results with a retrained YOLO model on a public handle-specific dataset.

\subsection{Runtime Analysis}
\input{tables/compute_analysis}
\tabref{tab:compute} decomposes the runtime of the system into individual components.
While the agent executes more navigation subpolicy calls (this includes driving to objects to open), each manipulation takes longer on average, resulting in similar total times spent in each. We also find that the high-level reasoning takes up a significant fraction of time. We extend this with data from the simulation experiments to be able to break it down into components.
In particular the LLM queries for high-level reasoning take up the majority of this time. This demonstrates the importance of current work for compact and fast inference, which is currently receiving a lot of attention~\citeS{chavan2024faster}, as well as the importance of open-source models~\citeS{touvron2023llama} that can be run locally instead of purely through an API.

While the current implementation of the scene graph is not optimized for speed, and the graph is fully recomputed at each time step instead of only updating areas that received new observations, we find its overall time impact to be reasonable, as it is only required at high-level reasoning steps.

\subsection{Extended Future Work Discussion}
In this work, we introduce scene graphs as an efficient and scalable representation for high-level, language model based reasoning, by encoding the scene graphs in a structured language representation and the incorporation of knowledge about distances and unexplored areas.

In future work, we aim to relax the assumptions about accurate perception, fully constructing scene graphs from noisy sensor inputs, as e.g. done in Hydra~\citeS{hughes2022hydra} or the direct incorporation of open-vocabulary representations~\citeS{werby23hovsg}.
While we currently encoding distances and spatial arrangements as adjectives and room-object relations, full maps provide much more dense spatial and geometric information. Research for more direct incorporation of such information, e.g. through vision-language models is very promising. Furthermore, more holistic approaches to incorporate spatial and semantic details in room clustering and classifications will be important to address non-standard layouts and designs.
Lastly, methods to incorporate more detailed visual feedback for the identification of object states and failure reasons are an important are to increase robustness and success over long tasks.

{\footnotesize
\bibliographystyleS{IEEEtran}
\bibliographyS{bibliography}
}

%% file: figures/tex-figures/real-world-map.tex
\setlength{\tabcolsep}{1pt}

\begin{figure}[t]
	\centering
	\resizebox{\linewidth}{!}{%
 \includegraphics[width=\textwidth,trim={0cm 0cm 0cm 0cm},clip,angle =0,valign=c]{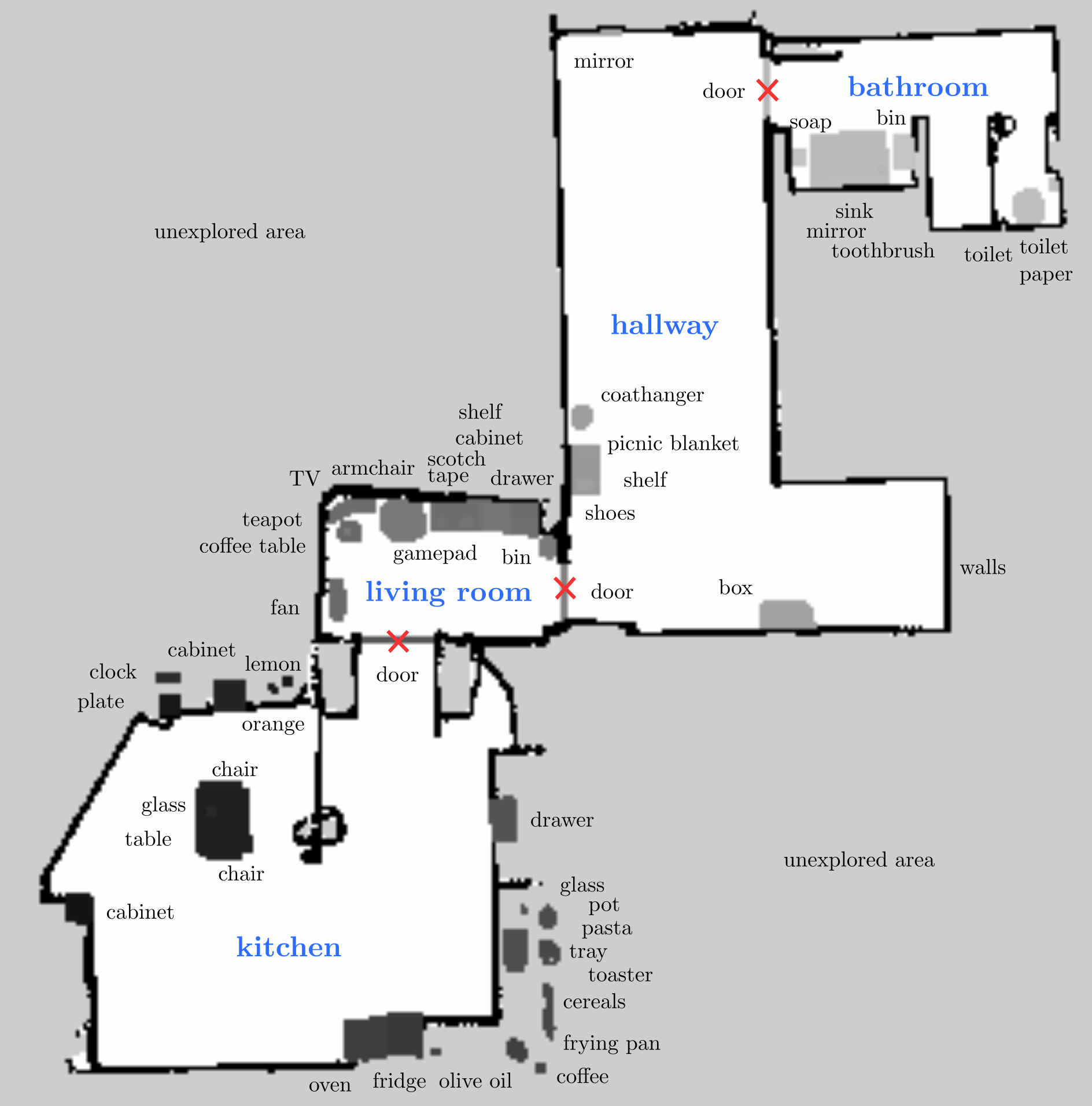}}

     \caption{Map of the real-world environment. Different intensities represent different object instances in the environment. Room annotations are for illustration and not used by our model. Object labels denote approximate object positions.}
  	\label{fig:real-world-map}
\end{figure}
\setlength{\tabcolsep}{6pt}

%% file: tables/scene_graph_structure.tex
\begin{table*}
    \centering
    \caption{Scene Graph Structure}
    \label{tab:scene_graph_structure}
    \renewcommand{\arraystretch}{1.3}
    \begin{threeparttable}
    \begin{tabularx}{\linewidth}{cl}
      \toprule
       Component & Description \\
      \midrule
       $\mathcal{G}_{\mathcal{V}} = (\mathcal{V},\mathcal{E})$
 & Navigational Voronoi graph that undergoes sparsification and covers traversable map areas \\
        & \\
        $\downarrow$ & \\
         & \\
       $\mathcal{G}_{\mathcal{V}}^{\mathcal{R}} = (\mathcal{V}', \mathcal{E}')$ with 
       $\mathcal{V}'=\{\mathcal{V}_1, \mathcal{V}_2, \ldots, \mathcal{V}_k \}$ & Deletion of edges with sufficient edge integrals computed over $\rho_{\mathcal{N}}$ (Eq. 2) results \\ 
       where $\mathcal{V}_i \cap \mathcal{V}_j = \emptyset$ for all $i,j \in \{1,\ldots, k\}$ with $i \neq j$ & in pair-wise edge-disjoint sets $\mathcal{V}_i$ with each set covering a room $r_i$. \\
        & \\
        $\downarrow$ & \\
         & \\
         & We define the actual scene graph $\mathcal{G}_{S}$ over all object nodes $\mathcal{V}_{O}$ and room supernodes $\mathcal{V}_{R}$. \\
        & Each room $r_i \in \mathcal{V}_{R}$ holds edges its corresponding Voronoi nodes $\mathcal{V}_i \in \mathcal{V}'$. \\
        $\mathcal{G}_{S} = (\mathcal{V}_{O} \cup \mathcal{V}_\mathcal{R}\,,\, \mathcal{E}_{OR} \cup \mathcal{E}_{RR})$ & Each room $r_i \in \mathcal{V}_{R}$ (and its Voronoi nodes $\mathcal{V}_{i }$) undergoes semantic classification (Fig. 3).  \\
         & The edges of $\mathcal{E}_{OR}$ connect objects $o \in \mathcal{G}_{S}$ with their associated room $r_i \in \mathcal{V}_{R}$ via Eq. 3. \\
        & $\mathcal{E}_{RR}$ are the edges connecting neighboring rooms $r_i \in \mathcal{V}_{R}$ to one another. \\
      \bottomrule
    \end{tabularx}
    \begin{tablenotes}[para,flushleft]
       \footnotesize      
       
     \end{tablenotes}
   \end{threeparttable}
\end{table*}

%% file: tables/distance_encoding.tex
    \begin{table}
        \centering
        \caption{Mapping of distances to natural language.}
        \label{tab:distance_encoding}
        \begin{tabularx}{\linewidth}{Y|Y}
          \toprule
            $\le$ Distance & Encoding \\
          \midrule
            \phantom{0}3.0 & very close\\
            10.0 & near \\
            20.0 & far \\
            \phantom{0}$\infty$ & distant\\
          \bottomrule
        \end{tabularx}
    \end{table}

%% file: figures/tex-figures/json-policy.tex
\setlength{\tabcolsep}{1pt}
\begin{figure*}[ht]
	\centering
	\resizebox{\linewidth}{!}{%
 \includegraphics[width=\textwidth,trim={0cm 0cm 0cm 0cm},clip,angle =0,valign=c]{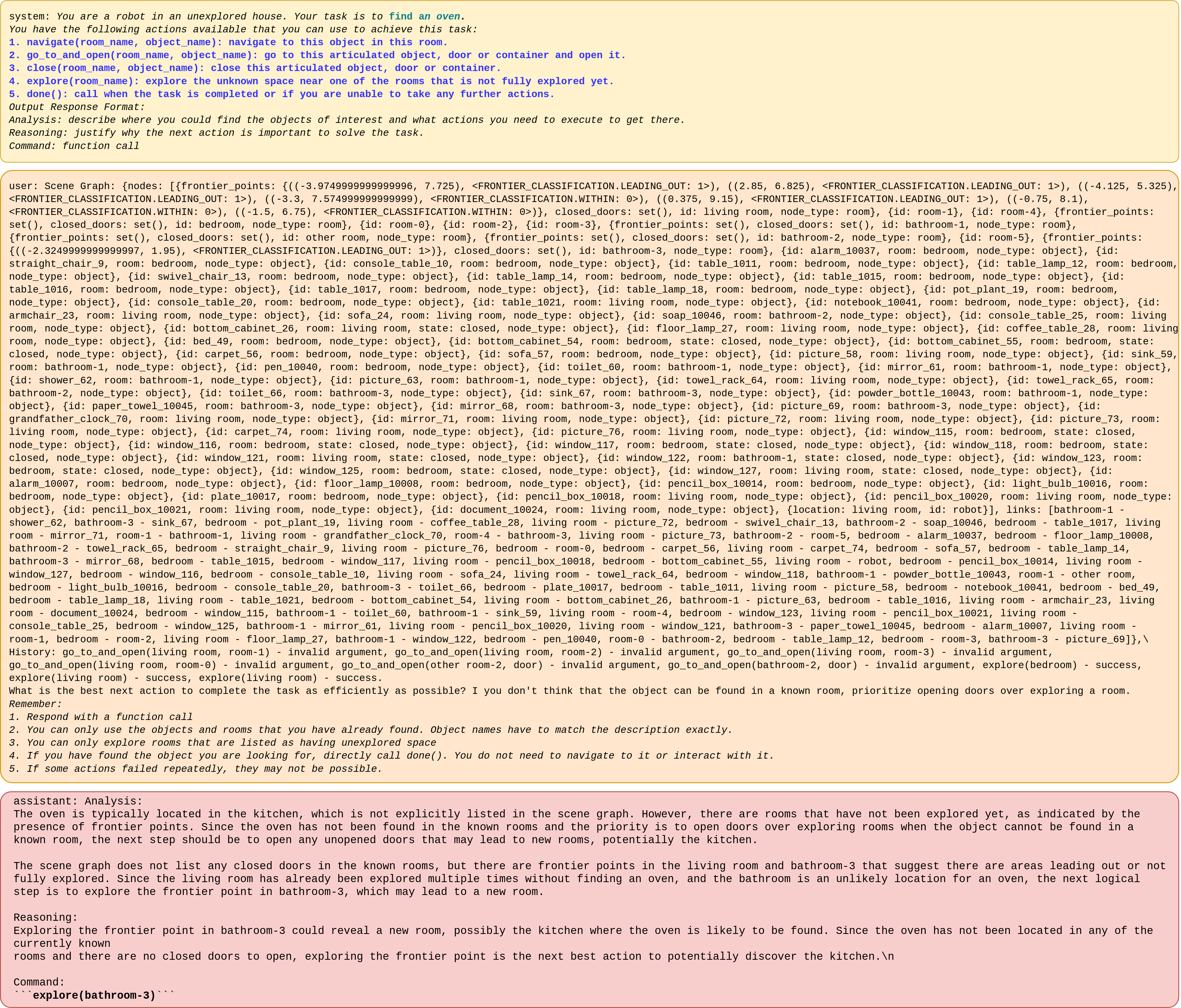}}
     \caption{Prompt of the Unstructured-LLM baseline. It receives the same instructions and notes as our approach, but instead of a structured knowledge encoding, it receives a pure JSON encoding of the scene-graph.}
  	\label{fig:json-policy}
\end{figure*}
\setlength{\tabcolsep}{6pt}

%% file: tables/closed_set_room_eval.tex
\begin{table}
    \centering
    \caption{Closed-Set Room Category Evaluation}
    \label{tab:room_cat_eval}
    \begin{tabularx}{0.8\linewidth}{lc|cc}
      \toprule
        \multirow{2}{*}{Environment} & \multirow{2}{*}{Open-Set} & \multirow{2}{*}{Purity $G_{\mathcal{V}}^{R}$} & Room Category  \\
         & &  & Accuracy \\
      \midrule
         iGibson & $\times$ & 0.615 & 0.276 \\
         Real-World & $\checkmark$ &- & 0.901 \\
      \bottomrule
    \end{tabularx}
\end{table}

%% file: figures/tex-figures/hydra-vs-momallm-sg.tex
\begin{figure*}[t]
	\centering
	\resizebox{1.0\linewidth}{!}{%
 \includegraphics[width=\linewidth]{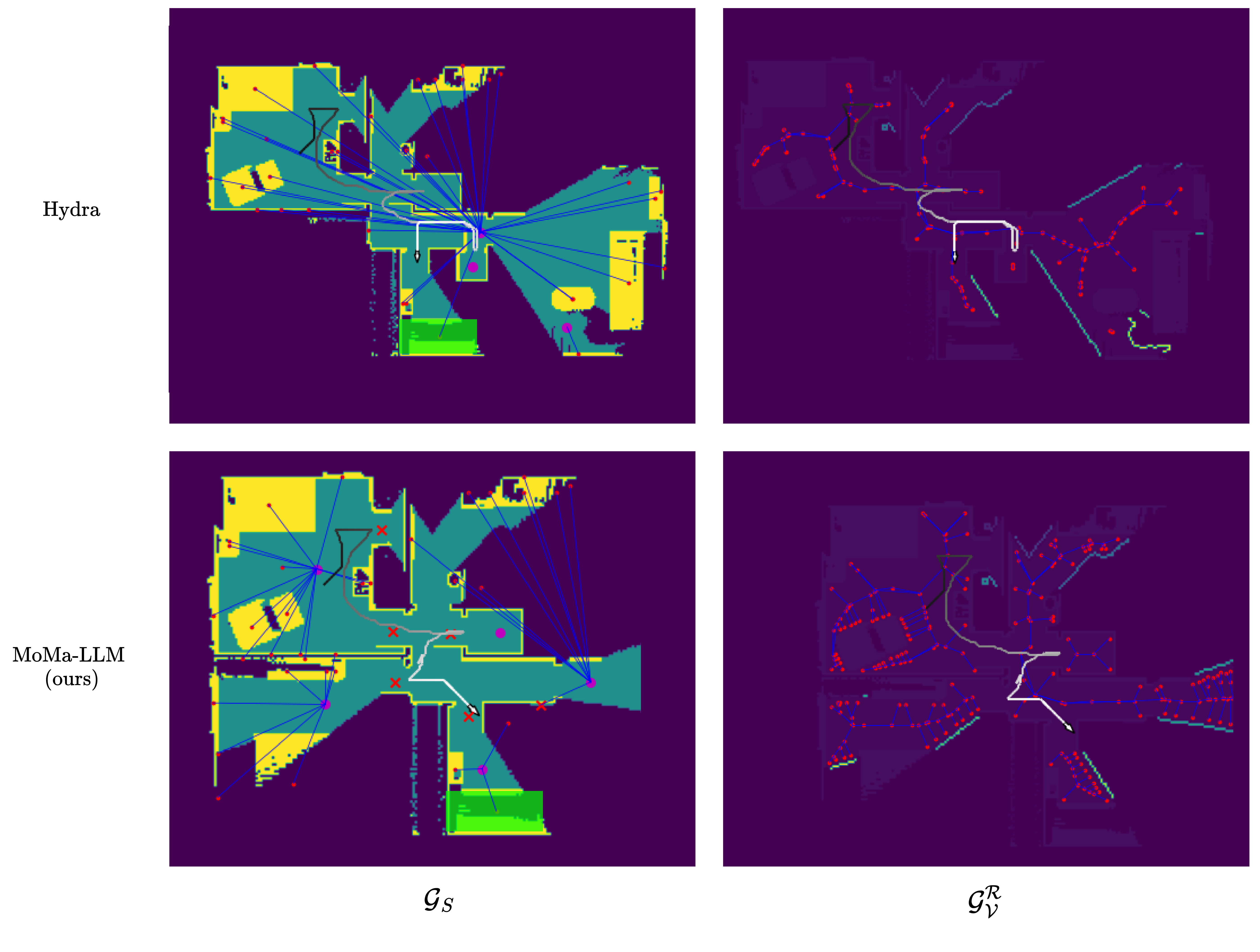}
 }
     \caption{Visualization of the scene graphs produced by Hydra compared to \ours{}. The left column represents the abstracted scene graphs $\mathcal{G}_{S}$ while the right column depicts the separated Voronoi graphs $\mathcal{G}_{\mathcal{V}}^{\mathcal{R}}$ as defined in \tabref{tab:scene_graph_structure}. The red crosses represent extracted door positions of \ours{}. The depicted scene is \textit{Merom\_1\_int} contained in the iGibson test scenes.} 
  	\label{fig:hydra-vs-momallm-sg}
\end{figure*}

%% file: figures/tex-figures/examples.tex
\setlength{\tabcolsep}{0pt}
\begin{figure*}[t]
	\centering
	\resizebox{\linewidth}{!}{%
 \begin{tabular}{c}
 \includegraphics[width=\linewidth,trim={0cm 0cm 0cm 0cm},clip,angle =0,valign=c]{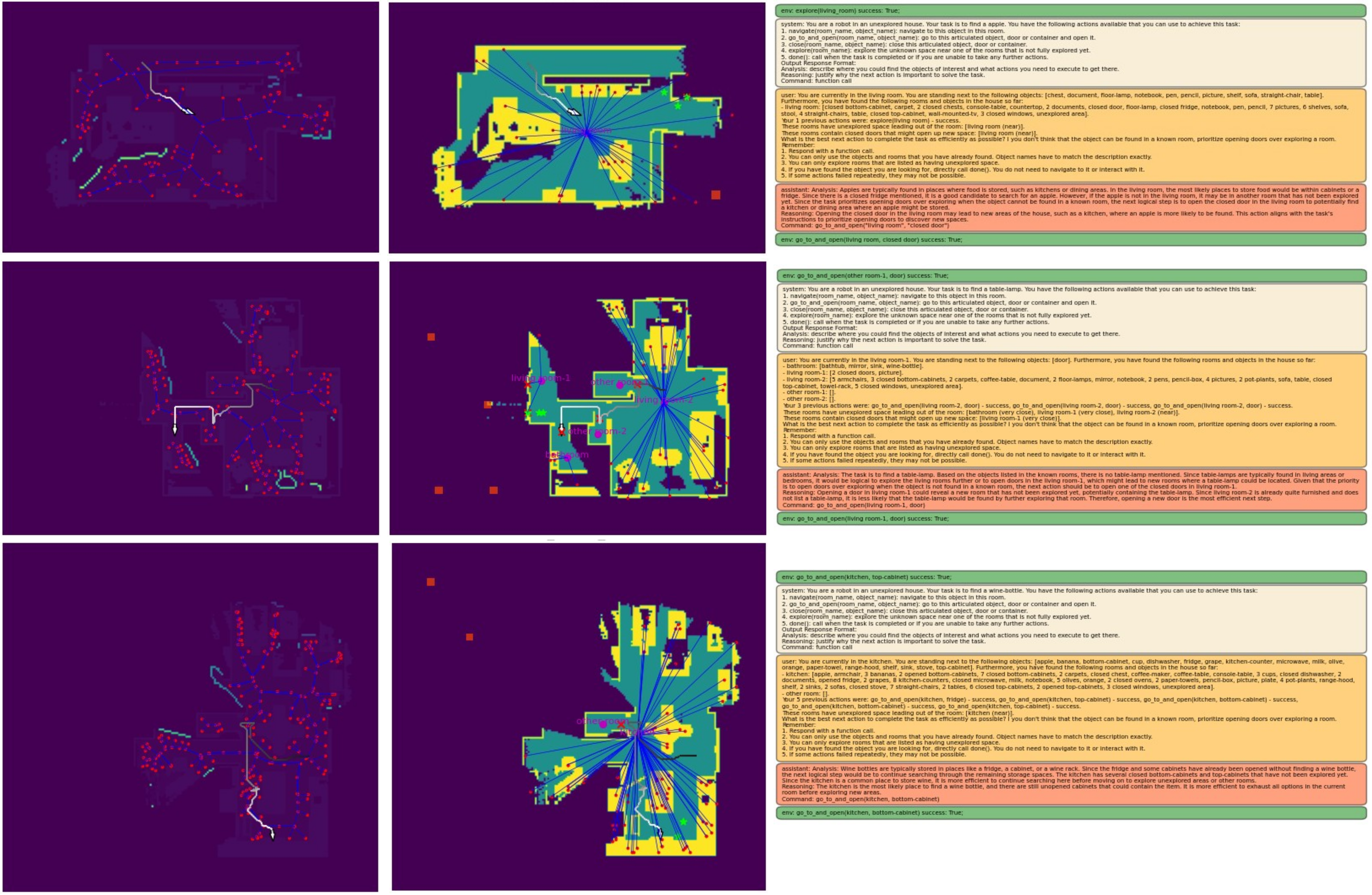} \\
 \end{tabular}}
	% \vspace{-0.2cm}
     \caption{Examples of the reasoning of Voronoi graph and identified frontiers (left), scene graph (middle), \ours{} (left). The green environment feedback is not provided to the LLM. Black-white: agent trace, red crosses: closed doors, red rectangles: undiscovered target objects, green stars: next selected navigation / interaction points. The bottom figure illustrates a subpotimal room clustering, which results in a very large room. We find the model to be robust against these clusterings.} 
  	\label{fig:examples}
\end{figure*}
\setlength{\tabcolsep}{6pt}

%% file: tables/fuzzy_search.tex
    \begin{table*}
        \scriptsize
        \centering
        \caption{Results on the fuzzy search tasks in the Real World. }
        \label{tab:fuzzy_search}%
        \begin{threeparttable}
        \begin{tabularx}{\linewidth}{l|cY}
          \toprule
            Task & Success & Reasoning \\
          \midrule
            I am hungry. Find me something for breakfast.      & \cmark & Found cereals, toast, coffee. Then opened fridge to look for milk or eggs. Found milk in the fridge and decided these are sufficient items.\\
            Find me something to wash my hands.                & \cmark & Searched kitchen for a sink. When not finding one, searches for storage room or bathroom until it finds the sink in the bathroom.\\
            I feel sleepy. Find me something to wake up.       & \cmark & The teapot is associated with tea that contains caffeine and can help someone wake up.\\
            Find things to set the kitchen table.              & \cmark & Explored kitchen, opening cabinets. Found knife, glasses, plates. Continued to look for forks or spoons. When not finding them, called done. (No forks or spoons existed).\\
            Find me the book in the living room.               & \cmark & Explored until finding living room, then opened cabinet looking for book, found it inside. \\
          \midrule
            Pour me a glass of milk.                           & \cmark & Finds milk in fridge, glass on table. Navigates between the two, assuming to transport the last object. Then terminates, reasoning that it has found and navigated to both.\\
            Turn on the oven.                                  & \cmark & Finds the oven and calls done(): "Turning on oven is implied as completion of the task".\\
            What's the time?                                   & \cmark & Finds the clock and calls done(): "The clock is the object that will provide the time".\\
          \bottomrule
        \end{tabularx}
        \begin{tablenotes}[para,flushleft]
           \footnotesize      
           Notes: Top: fuzzy search queries. Bottom: infeasible task queries. Success in these tasks is evaluated by human judgment as a reasonable response. The reasoning has been qualitatively paraphrased for brevity.
         \end{tablenotes}
       \end{threeparttable}
       \vspace{-0.3cm}
    \end{table*}

%% file: tables/compute_analysis.tex
\begin{table}
    \centering
    \caption{Evaluation of the average runtime per component.}
    \label{tab:compute}
    \begin{threeparttable}
    \begin{tabularx}{\linewidth}{cl|YYY}
      \toprule
        & Component & Time per step (\unit{\second}) & Total time (\unit{\second}) & Share (\%) \\
      \midrule
        \parbox[t]{1mm}{\multirow{5}{*}{\rotatebox[origin=c]{90}{Real-World}}}
        & \textit{Total}                     & 24.1                & 457                & 100\\
        % \midrule
        & \textit{Low-level Execution}       & 13.2                & 250                & \phantom{0}55\\
        & Navigation                         & \phantom{0}7.0      & 139                & \phantom{0}30\\
        & Manipulation                       & 13.9                & 111                & \phantom{0}24\\
        % \midrule
        & \textit{High-level Reasoning}  & 10.9      & 207                & \phantom{0}45\\
        \midrule
        \parbox[t]{1mm}{\multirow{4}{*}{\rotatebox[origin=c]{90}{Simulation}}}
        & \textit{High-level Reasoning}& 11.1              & \phantom{0}91.1             & \phantom{0}- \\
        & Scene Graph Construction           & \phantom{0}3.0    & \phantom{0}31.5             & \phantom{0}- \\
        & Room Classification                & \phantom{0}0.6    & \phantom{00}5.0   & \phantom{0}- \\
        & LLM Reasoning                      & \phantom{0}7.5    & \phantom{0}64.6             & \phantom{0}- \\
      \bottomrule
    \end{tabularx}
    \begin{tablenotes}[para,flushleft]
       \footnotesize      
       Notes: Time per step is the average time for one high-level step or subpolicy call. Total time is the average total time per episode. Real-world numbers are averaged over three episodes of the fuzzy-search experiments. Simulation numbers are averaged over 175 episodes in the iGibson simulator.
     \end{tablenotes}
   \end{threeparttable}
\end{table}